\documentclass[times,final]{elsarticle}

\usepackage{framed,multirow}

\usepackage{amssymb}
\usepackage{latexsym}
\usepackage{bm}
\usepackage{amsmath}
\usepackage[labelfont=bf,textfont=md]{caption}
\usepackage{graphicx}
\usepackage[utf8]{inputenc}
\usepackage{tikz}
\usepackage[ruled,vlined]{algorithm2e}
\usepackage{algorithmic}
\usepackage[utf8]{inputenc}
\usepackage[english]{babel}
\usepackage{amsthm}
\usepackage{amsmath}

\usepackage[lofdepth,lotdepth]{subfig}

\usepackage[hidelinks]{hyperref}
\hypersetup{
  colorlinks   = false, 
  urlcolor     = blue, 
  linkcolor    = green, 
  citecolor   = red 
  }

\graphicspath{ {./images/} }

\usetikzlibrary{shapes.geometric, arrows}
\tikzstyle{startstop} = [rectangle, rounded corners, minimum width=3cm, minimum height=1cm,text centered, draw=black, fill=red!30]
\tikzstyle{io} = [trapezium, trapezium left angle=70, trapezium right angle=110, minimum width=3cm, minimum height=1cm, text centered, draw=black, fill=blue!30]
\tikzstyle{process} = [rectangle, minimum width=3cm, minimum height=1cm, text centered, draw=black, fill=orange!30]
\tikzstyle{decision} = [diamond, minimum width=3cm, minimum height=1cm, text centered, draw=black, fill=green!30]
\tikzstyle{arrow} = [thick,->,>=stealth]

\theoremstyle{definition}

\theoremstyle{remark}

\usepackage{url}
\usepackage{xcolor}
\definecolor{newcolor}{rgb}{.8,.349,.1}

\begin{document}


\begin{frontmatter}

\title{PAGP: A physics-assisted Gaussian process framework with active learning for forward and inverse problems of partial differential equations\tnoteref{tnote1}}



\author[1]{Jiahao {Zhang}\corref{cor1}}

\author[1]{Shiqi {Zhang}\corref{cor1}}

\author[1,2]{Guang {Lin}\corref{cor2}}

\cortext[cor1]{These authors contributed equally.}

\cortext[cor2]{Corresponding author. 
E-mail: guanglin@purdue.edu.}

\address[1]{Department of Mathematics, Purdue University, West Lafayette, IN 47906, USA}
\address[2]{School of Mechanical Engineering, Department of Statistics (Courtesy), Department of Earth, Atmospheric, and Planetary Sciences (Courtesy), Purdue University, West Lafayette, IN 47907, USA}



\begin{abstract}
In this work, a Gaussian process regression(GPR) model incorporated with given physical information in partial differential equations(PDEs) is developed: physics-assisted Gaussian processes(PAGP). The targets of this model can be divided into two types of problem: finding solutions or discovering unknown coefficients of given PDEs with initial and boundary conditions. We introduce three different models: continuous time, discrete time and hybrid models. The given physical information is integrated into Gaussian process model through our designed GP loss functions. Three types of loss function are provided in this paper based on two different approaches to train the standard GP model. The first part of the paper introduces the continuous time model which treats temporal domain the same as spatial domain. The unknown coefficients in given PDEs can be jointly learned with GP hyper-parameters by minimizing the designed loss function. In the discrete time models, we first choose a time discretization scheme to discretize the temporal domain. Then the PAGP model is applied at each time step together with the scheme to approximate PDE solutions at given test points of final time. To discover unknown coefficients in this setting, observations at two specific time are needed and a mixed mean square error function is constructed to obtain the optimal coefficients. In the last part, a novel hybrid model combining the continuous and discrete time models is presented. It merges the flexibility of continuous time model and the accuracy of the discrete time model. The performance of choosing different models with different GP loss functions is also discussed. The effectiveness of the proposed PAGP methods is illustrated in our numerical section.

\end{abstract}

\end{frontmatter}


\section{Introduction}
Nowadays data-driven machine learning(ML) models have achieved great success in scientific computing and discoveries across many disciplines \cite{PMLAI, MLTPP, ICDCNN, DL}. However, using ML models merely as black box functions might lead to poor performance due to ignoring the existing physical laws or other domain expertise. Also, most of current black box ML models often have large data requirements and limited generalization properties. Thus, combining the ML models with the governing physical laws which often takes the form of partial differential equations(PDEs) becomes a natural popular topic. There is already a vast amount of works in this area, including \cite{APIK, GPRCBVP, ISDE, PINN, MTFMPRGP}. For a more comprehensive review of the previous works, the readers are  referred to \cite{IPBMML}. 
Among all the data-driven ML models, Gaussian process regression(GPR), also known as Kriging in geostatistics, is a widely used non-parametric Bayesian model for constructing a cheap surrogate for complex science and engineering problems. Gaussian process is uniquely determined by its prescribed forms of mean and covariance functions. It has a probabilistic workflow which enjoys analytical tractability and returns robust variance estimates from its posterior distribution. This also naturally quantifies the uncertainties of the model. See Section \ref{GPR} for a more detailed introduction of GPR. In this paper, we aim at integrating physical information contained in PDEs with GPR model to solve both the forward problem, i.e. finding the solutions of given PDEs, and the inverse problem, i.e. discovering the unknown coefficients of given PDEs. Two previous works of dealing with this kind of problems are briefly mentioned in the following.

In \cite{MLLDE} and \cite{NumGP}, the authors proposed a machine learning model called physics-informed GP(PIGP) to integrate the physical laws, such as conservation of mass, momentum and energy which are expressed by PDEs, with the standard GP. PIGP can be divided into two submodels depending on whether the temporal discretization is performed. A review and compare of PIGP and physics-informed neural network(PINN \cite{PINN}) models can be found in \cite{PIGP}. Consider the linear partial differential equations of the following form,
$$
\mathcal{L}_{\mathbf{x}, t}^c u = 0, x \in \Omega, t \in [0, T]
$$
where $c$ is a set of coefficients, $\mathcal{L}_{\bold{x}, t}^c$ is a linear operator of $\bold{x}$ and $t$ on $u$ with coefficients $c$. The continuous PIGP model first assumes that $u$ is a mean zero GP, i.e. $u((\bold{x},t); \bold{\theta}) \sim \mathcal{GP}(0, k_u((\bold{x},t), (\bold{x}',t'); \bold{\theta}))$. Because the linear transformation of a GP is still a GP \cite{LOSPDEGPR}, $U((\bold{x}, t); \bold{\theta}) := \mathcal{L}_{\bold{x},t}^c u((\bold{x}, t); \bold{\theta}) \sim \mathcal{GP}(0, k_U((\bold{x},t), (\bold{x}',t'); \bold{\theta}))$ is also a GP which has the same hyper-parameters with $u$. The two GPs are correlated as $k_U((\bold{x},t), (\bold{x}',t'); \bold{\theta})) = \mathcal{L}_{\bold{x}, t}^c \mathcal{L}_{\bold{x'}, t'}^c k_u((\bold{x},t), (\bold{x}',t'); \bold{\theta}))$. And their shared hyper-parameters are jointly optimized in the training process of GP. In this way, the PDE constrains are incorporated into Gaussian process regression. PIGP can also handle the inverse problems as the coefficients $c$ can be learned as additional hyper-parameters of the GP. Note that if the PDE has nonlinear terms, they must be linearized first before performing the above process. For the discrete PIGP model, the spatial and temporal coordinates are treated differently instead of viewing $t$ as another dimension of $\bold{x}$. The temporal domain need to be discretizated by a pre-chosen scheme, such as backward Euler, Runge-Kutta, etc. Then, the desired PDE solution as a function of spatial variables is approximated by the GP surrogate and the physical principles are incorporated similarly into the GP as the continuous model. The inverse problem is also handled similarly by treating the unknown coefficients as additional hyper-parameters in the GP training process. PIGP provides a mathematically elegant way to integrate physical laws with GP. However, the continuous model cannot handle nonlinear operators which is very common in practical applications. As for the discrete model, independent GPs are needed to be trained in every time step, each of which has a computational complexity $O(n^3)$ due to the need to invert a $n \times n$ matrix. The matrix size $n$ is determined by the number of training data points.


Another method called physics-informed Kriging(PhIK) is presented by Xiu, et al.\cite{PhIK}. The physical principles are integrated into GP in a different way by designing specific mean and covariance functions. As we mentioned above, GP is uniquely determined by its assumed mean and covariance functions. The unknown hyper-parameters are estimated from the data in the training process. In PhIK, the authors first compute the mean and covariance functions from realizations of available stochastic models, i.e. realizations of stochastic partial differential equation(SPDE) solutions. Then the predictions are obtained using the constructed mean and covariance functions. Thus there is no optimization process. PhIK allows predictions from data without the complex optimization problem in GP, but the accuracy of the model depends on the accuracy of the SPDEs. So, it is not suitable in situations where the SPDE solution is expensive to obtain or physical models are only partially known.

In this work, we propose a new way to incorporate physical principles into Gaussian process regression model. In a standard GP, the optimal hyper-parameters of the mean and covariance functions can be optimized in two different ways, i.e. minimizing the negative log marginal likelihood(NLML) or using the method of cross-validation. In PAGP models, three types of loss functions are constructed based on those two approaches. For the first method, we apply the leave-one-out cross-validation(LOO-CV) to construct a loss function. The log of validation density is used as the cross-validation measure of fit. An additional penalty term is added to this loss function following the idea of penalized GPR. This term is actually consist of the sum of absolute errors of PDE residue on a set of collocation points. The second loss function is similar except that the measure of fit for the LOO-CV is the square error. As for the last one, a same penalized term is added to the original NLML function. Moreover, an adaptive weight selection procedure is proposed to determine the weight coefficients multiplied to the penalized term in order for the loss function to be meaningful. The penalized term need to be computed on pre-set collocation points during the GP training process. Thus the derivatives of the GP predictions with respect to time $t$ and spatial locations $\bold{x}$ need to be derived first. As in PIGP, we develop continuous time models and discrete time models depending on different problem setting. For continuous time models, we follow the procedure in \cite{GPACLPL} and directly use the GP prediction formula to derive the analytic expression of the $n$-th order derivatives with respect to both $t$ and $\bold{x}$ according to given PDEs. See Section \ref{DGP} for more details. For discrete time models, the GP derivatives with respect to $\bold{x}$ can be calculated in a similar way. But the GP derivatives with respect to $t$ need to be computed differently. Here, the finite difference method is used to conduct this computation. Moreover, a novel two-step hybrid model which integrates continuous and discrete time models together is proposed. The first step follows the discrete time model but with a relatively big time step size. The predictions on each time step together with samples drawn from initial and boundary conditions of given PDEs build a training data set for the second step. Then the continuous time model can be utilized to obtain predictions on test points across the whole domain. See Section \ref{Hybrid} for a more detailed discussion.

Our objectives of this paper:
\begin{enumerate}
    \item Forward problem (Sections \ref{s231} and \ref{s241}): deriving solutions of partial differential equations with boundary conditions and initial conditions based on Gaussian process regression model under different problem settings. 
    
    \item Inverse problem (Sections \ref{s232} and \ref{s242}): finding the unknown coefficients in partial differential equations with data potentially contaminated with noises based on Gaussian process regression model. 
\end{enumerate}

Our contributions of this paper:
\begin{enumerate}
    \item A continuous time model (Section \ref{s23}) is developed for predicting solutions and estimating uncertainties of forward and inverse problems of PDEs. This model is concise and flexible while achieving relatively good accuracy.
    
    \item A discrete time model (Section \ref{s24}) is developed for forward and inverse problems of PDEs. This model is more accurate with respect to relative $L^2$ error and it can be used with a Bayesian active learning scheme to improve model performance. 
    \item A hybrid model (Section \ref{Hybrid}) is developed for forward problems of PDEs. It combines the above two different models and can effectively reduces the PDE residue error. 
    
    \item We introduce a new mechanism (Section \ref{s231}) for regularizing the GP training process effectively in small data regime.
    
    \item We put forth an active learning framework (Section \ref{AC}) that enables the synergistic combination of mathematical models and data to reduce the model uncertainties.
\end{enumerate}

The paper is organized as follows. In Sections \ref{GPR} and \ref{DGP}, we give a brief introduction to the famous Gaussian process regression(GPR) model and the derivatives of its
posterior mean. Then the continuous time, discrete time and hybrid PAGP models are constructed in Sections \ref{s23}, \ref{s24} and \ref{Hybrid}. An active learning scheme is proposed in Section \ref{AC}. In Section \ref{NR}, five numerical examples are presented to illustrate the performance of proposed models corresponding to different problem settings. Conclusions and future works are provided in Section \ref{Summary}.

\section{Methodology} \label{Method}
The goal of this paper is to solve the forward and inverse problems of partial differential equations using GP incorporated with physical information as the building block. In this work, we consider the parametrized partial differential equations of the general form,
\begin{equation} \label{e1}
   u_t - \mathcal{T}_{ \bold{x}}^{\lambda} u = 0, \bold{x} \in \Omega, t \in [0, T]  
\end{equation}
with boundary condition,
\begin{equation} \label{e2}
    u(\bold{x}, t) = g(\bold{x}, t), \bold{x} \in \Gamma, t \in [0, T]
\end{equation}
and initial condition,
\begin{equation}\label{e3}
   u(\bold{x}, 0) = h(\bold{x}), \bold{x} \in \Omega
\end{equation}
where $\mathcal{T}_{ \bold{x}}^{\lambda}$ is a general differential operator which can be linear or nonlinear. The subscript denotes the spatial location $\bold{x}$ which operator $\mathcal{T}$ acts on. The superscript denotes parameters $\lambda$ which can be only partially known. $\Omega$ is a subset of $\mathbb{R}^d$ and $\Gamma$ is the boundary of $\Omega$. For example, consider the one dimensional heat equation: $u_t - \lambda u_{xx} = 0$. Here, the differential operator is $\mathcal{T}_{ \bold{x}}^{\lambda} = \lambda \frac{\partial^2}{\partial x^2}$ and $\lambda$ is an unknown parameter. The forward problem we considered in this setting is to find the solution $u(\bold{x}, t)$ given specific boundary conditions $g(\bold{x}, t)$ and initial conditions $h(\bold{x})$ and coefficients $\lambda$, while the inverse problem is to recover unknown coefficients $\lambda$ in the PDEs. Note that for both types of problem, the boundary and initial conditions of the PDEs can also be replaced by observations potentially contaminated with noises.

\subsection{Gaussian process regression} \label{GPR}
Gaussian process regression is a popular non-parametric Bayesian model. It enables us to build a surrogate with small data set. Let's suppose there is an unknown function $f$ which needs to be approximated:
$$y = f(\mathbf{x})$$ where $\mathbf{x} \in \mathbb{R}^d$. 
We denote the observation set to be $D = \{\mathbf{x}_i, y_i\}_{i=1}^n = (\mathbf{X}, \mathbf{y})$. And $f(\mathbf{x})$  is assumed to be a zero mean GP, i.e. $f \sim \mathcal{GP}(\mathbf{f}|\mathbf{0}, k(\mathbf{x}, \mathbf{x}';\mathbf{\theta})) $, where $k$ is the Gaussian process covariance function and $\mathbf{\theta}$ is a set of corresponding hyper-parameters. This assumption essentially reflects our prior belief about the unknown function $f$. 

If we also assume a Gaussian likelihood, the optimal hyper-parameters in the covariance function can be obtained by minimizing the negative log marginal likelihood of the model,
\begin{equation} \label{e4}
L_{NLML} := \mbox{log} p(\mathbf{y} | \mathbf{x}, \theta) = -\frac{1}{2} \mbox{log} |\mathbf{K}| -
\frac{1}{2} \mathbf{y}^T \mathbf{K}^{-1} \mathbf{y} - \frac{n}{2} \mbox{log} 2\pi
\end{equation}
where $\mathbf{K} = (K_{ij})_{i,j = 1}^n$ and $K_{ij} = k(\mathbf{x}_i,\mathbf{x}_j;\theta)$.

The posterior distribution of a GP is tractable and the prediction for a new output $f_*$ at a new input location $\mathbf{x}_*$ is given as 
$$
p(f_* | \mathbf{y}, \mathbf{X}, \mathbf{x}_*) = \mathcal{N}(f_*|\mu_*(\mathbf{x}_*), \sigma^2_*(\mathbf{x}_*))
$$

\begin{equation} \label{e5}
\mu_*(\mathbf{x}_*) = \mathbf{k}_{*n} \mathbf{K}^{-1} \mathbf{y}
\end{equation}

\begin{equation} \label{e6}
\sigma^2_*(\mathbf{x}_*) = \mathbf{k}_{**} - \mathbf{k}_{*n} \mathbf{K}^{-1} \mathbf{k}_{*n}^T
\end{equation}
where $\mathbf{k}_{**} = k(\mathbf{x}_*, \mathbf{x}_*)$ and $\mathbf{k}_{*n}  = [k(\mathbf{x}_*, \mathbf{x}_1), \cdots, k(\mathbf{x}_*, \mathbf{x}_n)]$. Equation \ref{e5} is the posterior mean $\mu_*(\mathbf{x}_*)$ of the GPR model and Equation \ref{e6} is the posterior variance which quantifies the uncertainty of the model naturally. Moreover, if there are noises in the observation set, we assume the noise to be Gaussian white noise with variance $\delta^2$. Then the resulting posterior mean and variance can be obtained by replacing the covariance matrix $K$ in Equations \ref{e5} and \ref{e6} by $K + \delta^2 I$.

Another method to find the optimal hyper-parameters of the covariance function is cross-validation(CV) which is always used in the $k$-fold cross-validation setting. We discuss an extreme case when $k = n$, where $n$ is the number of training points. This is also known as leave-one-out cross-validation(LOO-CV). To construct the loss function, two different measures of fit can be applied, i.e. the log of the validation density and the square error. The loss functions are the following when case $i$ is left out for the two different measures respectively,
\begin{equation} \label{e7}
    l_i = -\frac{1}{2}log \sigma_i^2 - \frac{(y_i - \mu_i)^2}{2 \sigma_i^2} - \frac{1}{2}log 2\pi
\end{equation}

\begin{equation} \label{e8}
    r_i = (y_i - \mu_i)^2
\end{equation}
where $\mu_i$ and $\sigma_i$ are computed by Equations \ref{e5} and \ref{e6} at location $x_i$. The training set $D_{-i} := D \setminus \{x_i, y_i\}$ is used to get the optimized hyper-parameters of covariance function in order to compute $\mu_i$ and $\sigma_i$. Accordingly, the LOO-CV loss function is the following for validation density measure,
\begin{equation}\label{e9}
    L_{\textit{LOO-VD}} := \sum_{i=1}^n l_i 
\end{equation}
For the square error measure, the LOO-CV loss function is, 
\begin{equation}\label{e10}
    L_{\textit{LOO-SE}} := \sum_{i=1}^n r_i 
\end{equation}
Now, the optimal hyper-parameters of the GP model can be computed by minimizing the above equations.

\subsection{Derivatives of Gaussian process regression} \label{DGP}
In order to incorporate the PDE constrains into GP model, we need to find a way to estimate different order of derivatives of the PDE solution. In our setting, we assume the PDE solution $u(\bold{x}, t)$ is a zero mean GP. According to Section \ref{GPR}, Equation \ref{e5} gives us the predictive mean of a GP at input $\bold{x}^*$. Following the ideas in \cite{GPACLPL}, we can derive the first and second order derivatives of the GP model as the corresponding derivatives of this mean function. The first order derivatives of predictive mean $\mu\left(\bold{x}^*\right)$ of the GP model at input $\bold{x}^* = (x_1^*, \cdots, x_t^*)$ with respect to $x_j^*$ are
\begin{equation} \label{e11}
    \frac{\partial \mu\left(\bold{x}^*\right)}{\partial x_{j}^*}=\sum_{l=1}^{n} \frac{\partial k\left(\bold{x}^*, \bold{x}_l\right)}{\partial x_{j}^{*}}\left[\mathbf{K}_{x x}^{-1}y\right]_{l} \quad \text { for } j=1, \ldots, n
\end{equation}
And the second order derivatives are
\begin{equation} \label{e12}
    \frac{\partial^2 \mu \left(\bold{x}^*\right)}{\partial x_{j}^*\partial x_{i}^*}=\sum_{l=1}^{n} \frac{\partial^2 k\left(\bold{x}^*, \bold{x}_l\right)}{\partial x_{j}^{*}\partial x_{i}^{*}}\left[\mathbf{K}_{x x}^{-1}y\right]_{l} \quad \text { for } i,j=1, \ldots, n
\end{equation}
The more general form of derivatives of GP model can be found in \cite{SGPRWD}.

\subsection{Continuous time model}
\label{s23}
\subsubsection{Forward problem: finding solutions of partial differential equations}\label{s231}
For the forward problem, assume the solution of Equation \ref{e1} is a zero mean Gaussian process with covariance function $k_u((\bold{x}, t), (\bold{x}', t'); \theta)$, i.e. $u(\bold{x}, t) \sim \mathcal{GP}(0, k_u)$, where $\theta$ is a set of hyper-parameters needed to be optimized. Given boundary conditions $g(\bold{x}, t)$ and initial condition $h(\bold{x})$, we can sample two set of training data corresponding to each kind of condition respectively, denoted by $D_b = \{ (\bold{x}_b, t_b), g_b(\bold{x}_b, t_b) \} $ for the boundary conditions and $D_0 = \{ (\bold{x}_0, 0), h(\bold{x}_0) \}$ for the initial condition. To incorporate the PDE constrains into GP model, we also need to sample a set of collocation points inside domain $\Omega$ denoted by $D_c = \{(\bold{x}_c, t_c)\}$. It should be noted that the PDE solutions $u(\bold{x}_c, t_c)$ are unknown at collocation points. So the training sets consist of $N_b$ number of points in $D_b$ and $N_0$ number of points in $D_0$. The number of constrain points is $N_c$. Suppose there is a set of test points at time $t = T$ to evaluate the performance of the model, which is denoted by $D_T = \{ (\bold{x}_T, T), u(\bold{x}_T, T) \}$. At this point, the hyper-parameter set $\theta$ must be trained using all the training data before meaningful predictions can be made. Three different loss functions are provided based on different methods and the corresponding performances are shown in the numerical example section. By adding a penalized term with a weight coefficient to the negative log marginal likelihood function (Equation \ref{e4}) or the LOO-CV loss function(Equations \ref{e9} and \ref{e10}), we construct the following loss functions $Loss_1$, $Loss_2$ and $Loss_3$:
\begin{equation} \label{e13}
    Loss_1 = L_{\textit{LOO-VD}} + \omega \textit{MSE}_r
\end{equation}

\begin{equation} \label{e14}
    Loss_2 = L_{\textit{LOO-SE}} + \omega \textit{MSE}_r
\end{equation}

\begin{equation}\label{e15}
    Loss_3 = L_{\textit{NLML}} + \omega \textit{MSE}_r
\end{equation}
where $L_{\textit{LOO-VD}}$, $L_{\textit{LOO-SE}}$ and $L_{\textit{NLML}}$ are computed using the training sets $D_b$ and $D_0$. $\textit{MSE}_r$ denotes the mean squared errors of PDE residue on collocation set $D_c$,
\begin{equation} \label{e16}
    \textit{MSE}_r = \frac{1}{N_c}\sum_{\bold{x}_c \in \Omega, t_c \in [0,T]} | u_t(\bold{x}_c, t_c) - \mathcal{T}_{ \bold{x}}^{\lambda} u(\bold{x}_c, t_c) |^2
\end{equation}
The weight coefficients $\omega$ need to be carefully adjusted to get the best performance. Here, an adaptive procedure is carried out to adjust the weights based on the relative magnitude of the two terms in the loss functions. We first set an initial value for $\omega$ and train the model. After the first training process, the weight coefficient is multiplied by a constant which we call it rate factor repeatedly until a pre-set maximal number. In this way, the weight coefficients can be gradually adjusted to better assist the training process.


Note that in Equation \ref{e16}, the differential operator $\mathcal{T}_{\bold{x}, t}^{\lambda}$ acts on solution $u$. This requires us to derive the corresponding orders of GP derivatives with respect to $\bold{x}$ and $t$, which is conducted following the procedure in Section \ref{DGP}. Then the GP loss function $Loss_1$, $Loss_2$ or $Loss_3$ is minimized with respect to hyper-parameter set $\theta$ given the PDE coefficients $\lambda$. Once the optimal hyper-parameters are computed, the posterior distribution mean and variance on test set $D_T$ can be obtained using Equation \ref{e5} and Equation \ref{e6}. In this way, the physical laws, i.e. PDE constrains, are integrated into GP model naturally and the full probabilistic workflow of GP is maintained.

\subsubsection{Inverse problem: discovering coefficients in partial differential equations}\label{s232}
The setting for inverse problem is the same as forward problem above except that there are some unknown coefficients in given PDEs, i.e. the physical laws are only partially known. In the continuous time model, we observe that the unknown coefficients can be treated as additional parameters similar to hyper-parameters of the GP covariance function. Accordingly, they can be jointly optimized by minimizing the GP loss function. The corresponding results of using loss function $Loss_1$, $Loss_2$ and $Loss_3$ are also discussed in the numerical experiments.

\subsection{Discrete time model}\label{s24}
\subsubsection{Forward problem: finding solutions of partial differential equations}\label{s241}
As in the continuous time model, suppose that the solution of given PDEs $u(\bold{x}, t) \sim \mathcal{GP}(0, k_u((\bold{x}, t), (\bold{x}', t'); \theta))$, where $\theta$ is the set of hyper-parameters of the covariance function. But unlike continuous time model, the spatial and temporal coordinates are treated differently. The PDEs in the problem need to be discretized along the temporal domain with a specific discretization scheme, such as Euler and Runge-Kutta methods, etc. For instance, if forward Euler scheme is applied to Equation \ref{e1}, we can obtain
$$
u^n = u^{n-1} + \Delta t \mathcal{T}_{\bold{x}}^{\lambda} u^{n-1}
$$
where the superscript denotes the number of time steps, e.g. $u^n(\bold{x}) = u(t_n, \bold{x})$ and $\Delta t$ is the time step size.

Given boundary conditions $g(\bold{x}, t)$(Equation \ref{e2}) and initial condition $h(\bold{x})$(Equation \ref{e3}), we can sample a set of data points inside the domain $\Omega$ at time $t_0 = 0$ from the initial condition $h(\bold{x})$. The derivatives of the solution $u$ involved in the time discretization scheme can also be obtained at $t_0 = 0$. Then, the time discretization scheme is utilized to derive training data points inside domain $\Omega$ at time $t_1 = \Delta t$, which are combined with samples drawn from the given boundary conditions $g(\bold{x}, t)$ to build a PAGP model at time $t_1$. Moreover, the predictions for the solution $u$ and its corresponding derivatives at given spatial locations can be obtained through the model. Now, the above two steps are repeated until the final time $t = T$. The last PAGP model can be used to make predictions for PDE solutions at desired time and test locations. For GP model at each time step, the physical information is involved through the constructed GP loss functions (Equations \ref{e13}, \ref{e14} and \ref{e15}). The derivatives of GP model with respect to spatial location $x$ are estimated using method in Section \ref{DGP} (Equations \ref{e11} and \ref{e12}) and the finite difference method is used to compute the derivatives of GP with respect to time $t$.
Note that if the initial conditions and boundary conditions are contaminated with some noises, then the resulting uncertainties will propagate along the time discretization scheme and we need to be more careful in this situation. 

\subsubsection{Inverse problem: discovering coefficients in partial differential equations}\label{s242}
For the inverse problem in discrete time model, we assume the solutions $u(\bold{x}, t; \lambda)$ containing unknown coefficients $\lambda$ can be observed at two specific time $0 < t_1 < t_2$, denoted respectively by $u_1(\bold{x}_1, t_1; \lambda)$ and $u_2(\bold{x}_2, t_2; \lambda)$. Using the standard GPR model, the predictions at $(\bold{x}_2, t_2)$ can be made using the samples $u_1(\bold{x}_1, t_1; \lambda)$, denoted by $\hat{u}_2(\bold{x}_2, t_2; \lambda)$. Similarly, the predictions at $(\bold{x}_1, t_1)$ can be made using the samples $u_2(\bold{x}_2, t_2; \lambda)$, denoted by $\hat{u}_1(\bold{x}_1, t_1; \lambda)$. Then, a mixed error function of the unknown coefficients $\lambda$ can be built as,
\begin{equation} \label{e17}
    E(\lambda) = \frac{1}{N_1}\sum_{\bold{x}_1 \in \Omega, t = t_1} (\hat{u}_1(\bold{x}_1, t_1; \lambda) - u_1(\bold{x}_1, t_1; \lambda))^2 + \frac{1}{N_2}\sum_{\bold{x}_2 \in \Omega, t = t_2} (\hat{u}_2(\bold{x}_2, t_2; \lambda) -u_2(\bold{x}_2, t_2; \lambda))^2
\end{equation}
where $N_1$ is the number of samples at $t=t_1$ and $N_2$ is the number of samples at $t=t_2$.

At this point, the error function $E(\lambda)$ is minimized to obtain the optimal values of $\lambda$, which is the best fit of coefficients for the given partial different equations.

\subsection{Hybrid model} \label{Hybrid}
In this section, we introduce our hybrid model which is a two-step process built on the continuous and discrete time models. Only forward problem is considered under this setting. Firstly, given PDEs are discretized along the temporal domain with a relatively big time step size. Then the discrete time model is applied to obtain predictions at each time step which we call them coarse layers in the numerical section. In the second step, the above predictions along with samples from given boundary and initial conditions can be combined together to form a training data set in this step. Then predictions can be made on any test point based on the continuous time model. The PDE constrains are involved the same way as in the training processes of the continuous and discrete time models through the built GP loss functions. 

The idea of hybrid model is very intuitive. In the first step, we use the discrete time model to construct training data inside domain $\Omega$. The time step size and the number of training points can be manually adjusted. The combination of coarse layer predictions and samples from boundary and initial conditions can effectively reduces the model uncertainties in the next step. Then the continuous time model is utilized. With the designed training data inside domain $\Omega$, we expect the predictive means are more accurate and variances are much less than that in the pure continuous time model. Also, it should be more computationally efficient as the time step size is bigger. In order to illustrate the performance of hybrid model, the test points can be randomly sampled across the whole domain $\Omega \times [0, T]$. We also present the mean absolute PDE residue errors on those test points in the numerical experiments for hybrid model to show that it can reduce the PDE residue errors effectively.

\subsection{Active learning}\label{AC}
Active learning is also known as optimal experimental design or sequential design in statistic literature. It aims at maximizing information acquisition with relatively small data set. To be more specific, the model is repeatedly updated by the data obtained from the experiments so that it is gradually improved. More recent works on this topic includes \cite{AL1, AL3, AL2, AL4}. In the problem setting of this paper, the training data set $D$ come from given boundary and initial conditions of given PDEs. To make accurate inference with relatively small data set, additional data points to augment the original observations can be efficiently determined by a data selection procedure, i.e. the active learning process. Here, the candidate pool can be viewed as the whole domain. This Bayesian active learning process can greatly reduce the model uncertainties under limited budget.

Assuming the training data set $D$ consists of $N$ samples. This represents the current state of knowledge, the most informative sample in the given PDE domain is picked by maximizing an acquisition function $a_N(x)$,
\begin{equation} \label{e18}
    \mathbf{x}_{N+1} = \textit{argmax}_{\mathbf{x} \in D} a_N(\mathbf{x})
\end{equation}
The acquisition function actually quantifies how much information we can get to evaluate or perform an expensive experiment at this data site. Then $(x_{N+1}, y_{N+1})$ is added to the original training data set $D$. At this point, the process stops if a pre-set criterion is achieved. Otherwise, the process repeats iteratively until it satisfies the stopping criterion or reaches the maximal number of iteration times. In our PAGP models, the acquisition function is chosen to be the variance of the posterior distribution:
\begin{equation}
a_N(\mathbf{x}) = \sigma^2_*(\mathbf{x})
\end{equation}
which quantifies how much uncertainties the model has for the current predictions. 

As for the stopping criterion, the active learning process stops if the relative $L^2$ error between the ground truth solutions of the PDEs and PAGP predictions on the test set is less than a chosen small value $\eta$.


\section{Numerical Results}\label{NR}
In this section, we present five numerical examples to illustrate the performance of our proposed models. Both the forward and inverse problems are considered in continuous and discrete time models, while hybrid model only focuses on the forward problem. We only perform the active learning scheme in discrete time model and it can be applied similarly to the other two models. The kernel function for GP model is chosen according to specific problems. For heat and advection equations, the square exponential kernel with automatic relevance determination(ARD) \cite{ARD} is applied. For burger's equation, we select the neural network kernel\cite{GPML} following the guidance of \cite{NumGP}. The topic of choosing appropriate kernel for GP belongs to the model selection problem. There are numerous previous works with different methodologies, for instance \cite{BF, BIC}. The loss functions $Loss_1$(Equation \ref{e13}), $Loss_2$(Equation \ref{e14}) and $Loss_3$(Equation \ref{e15}) are all optimized using BFGS algorithm. Moreover, the relationships between the relative $L^2$ error of the posterior distribution means and the number of training points or the number of collocation points are investigated. For hybrid model examples, we also present the mean absolute PDE residue error on test points.

\subsection{One dimensional heat equation}
For the continuous time model, let's consider the following one dimensional heat equation as an illustration example,

\begin{equation}\label{e20}
    (\frac{\partial}{\partial t} - \lambda \frac{\partial^2}{\partial x^2})u(x, t) = 0, \hspace{2mm}x \in [-\frac{\pi}{2}, \frac{\pi}{2}], \hspace{2mm}t > 0
\end{equation}
where $\lambda = 1$ is the diffusivity constant of the heat equation. The boundary conditions are,
\begin{equation}
    u(\frac{\pi}{2}, t) = e^{- t}  
\end{equation}
\begin{equation}
    u(-\frac{\pi}{2}, t) = -e^{-t}
\end{equation}
and initial condition is,
\begin{equation}\label{e23}
    u(x, 0) = sin(x) 
\end{equation}
The true solution for this problem is $ u(x, t) =  e^{-t} sin(x)$. Next, we present the performance of the continuous time model under two different settings.

\begin{figure}[h]
\centering
\includegraphics[width=0.8\textwidth]{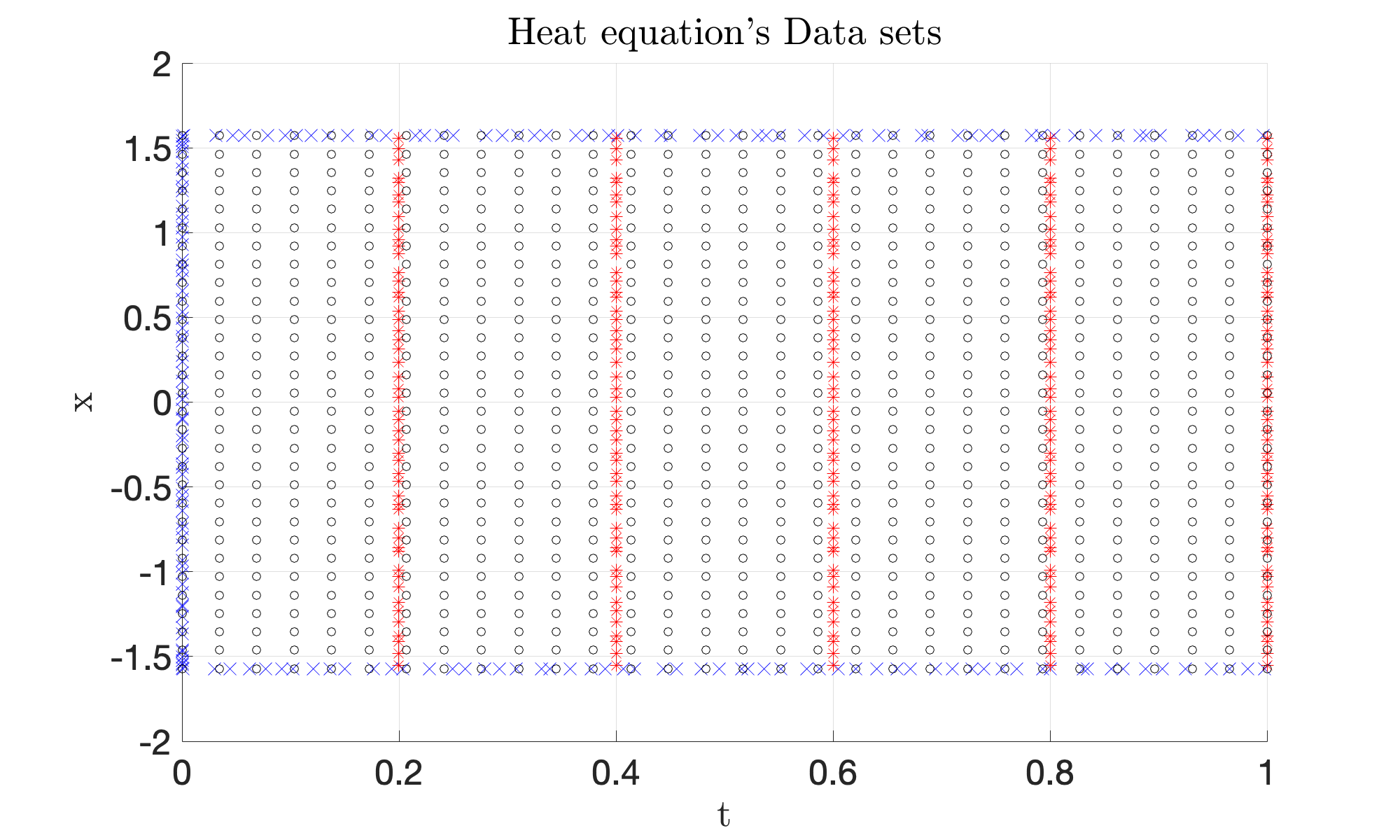}
\caption{Data sets of Equation \ref{e20}: training, collocation, and test data sets. The x axis denotes the time and the y axis denotes the spatial locations. The blue crosses are $180$ training data sampled from boundary and initial conditions. The red stars are $250$ test data at time $t = [0.2, 0.4, 0.6, 0.8, 1.0]$. The black circles are $900$ collocation points inside the domain.}
\label{fig:Sca_heat}
\end{figure}

For the forward problem, the coefficient $\lambda = 1$ is known. The boundary and initial conditions are given and we can sample a set of points as training data. We aim at finding solutions for Equation \ref{e20} at certain time $t$. According to our continuous time model in Section \ref{s23}, the temporal domain is treated the same as the spatial domain. So the number of input dimensions is two and physics-assisted Gaussian process is directly performed to make predictions at five times $t = [0.2, 0.4, 0.6, 0.8, 1.0]$s. The following square exponential kernel with ARD is applied,

\begin{equation}
    k((t, x), (t', x')) = \sigma^2 exp(-\frac{(t-t')^2}{2l_1} - \frac{(x-x')^2}{2l_2})
\end{equation}
where $\sigma$ and $l = ( l_1, l_2)$ are the hyper-parameters of this kernel function and they are jointly learned by minimizing the loss function $Loss_1$, $Loss_2$ or $Loss_3$ in Section \ref{s231}. The initial weight coefficients are set to be $\omega = 1$ for all thress loss functions. The rate factors are $1.2$ for $Loss_1$ or $Loss_2$ and $0.6$ for $Loss_3$. The maximal iteration number for updating the weight coefficients is $5$. Figure \ref{fig:Sca_heat} shows the training, collocation and test points for this example. The blue crosses are 180 training points sampled from boundary and initial conditions. The black circles are 900 collocation points inside the domain and the red stars are 250 test points. 

\begin{figure}[h]
\centering
\subfloat[a][]{
\includegraphics[width=0.45\textwidth]{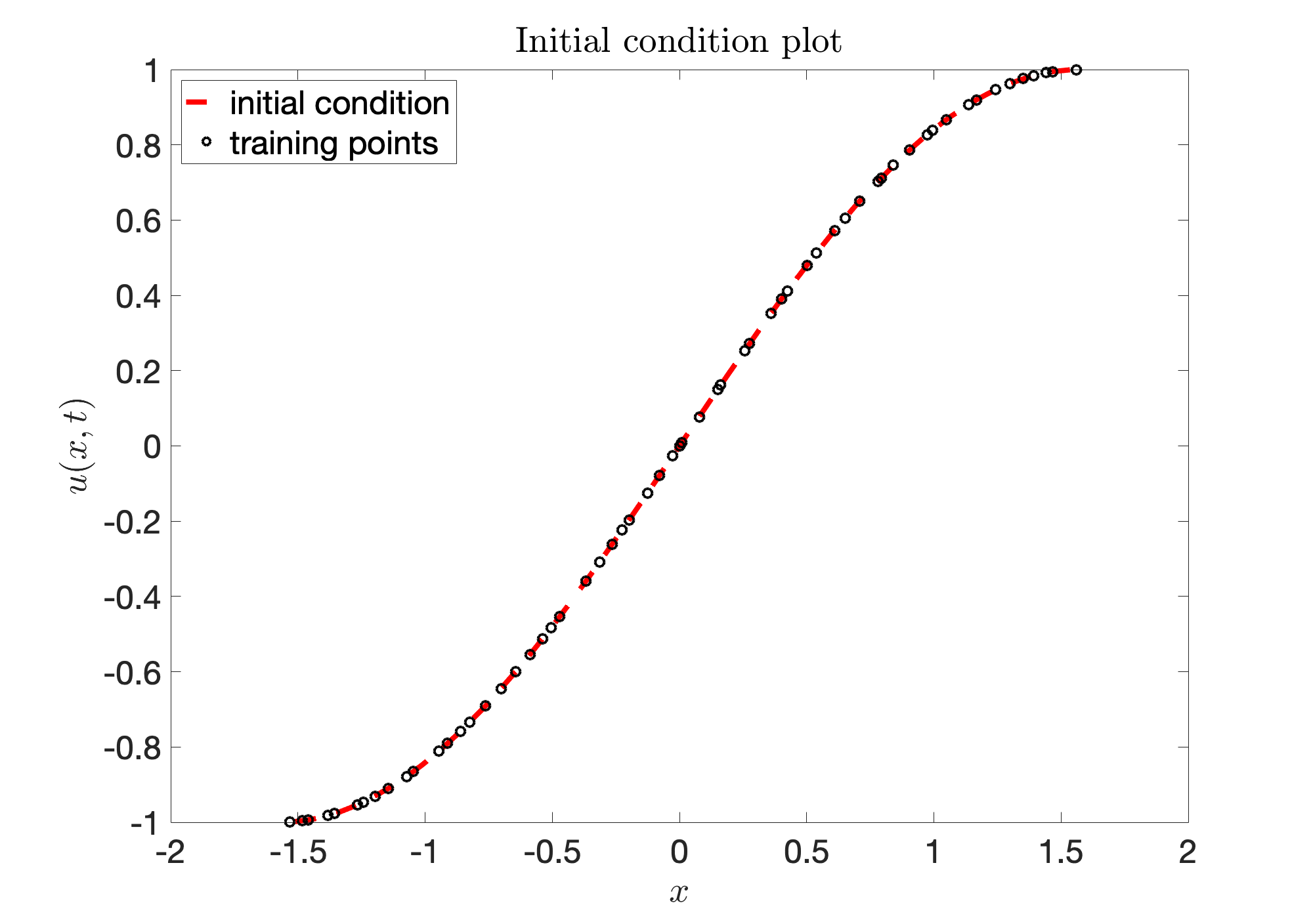}
\label{fig:Heat1d_predint}}
\qquad 
\subfloat[b][]{
\includegraphics[width=0.45\textwidth]{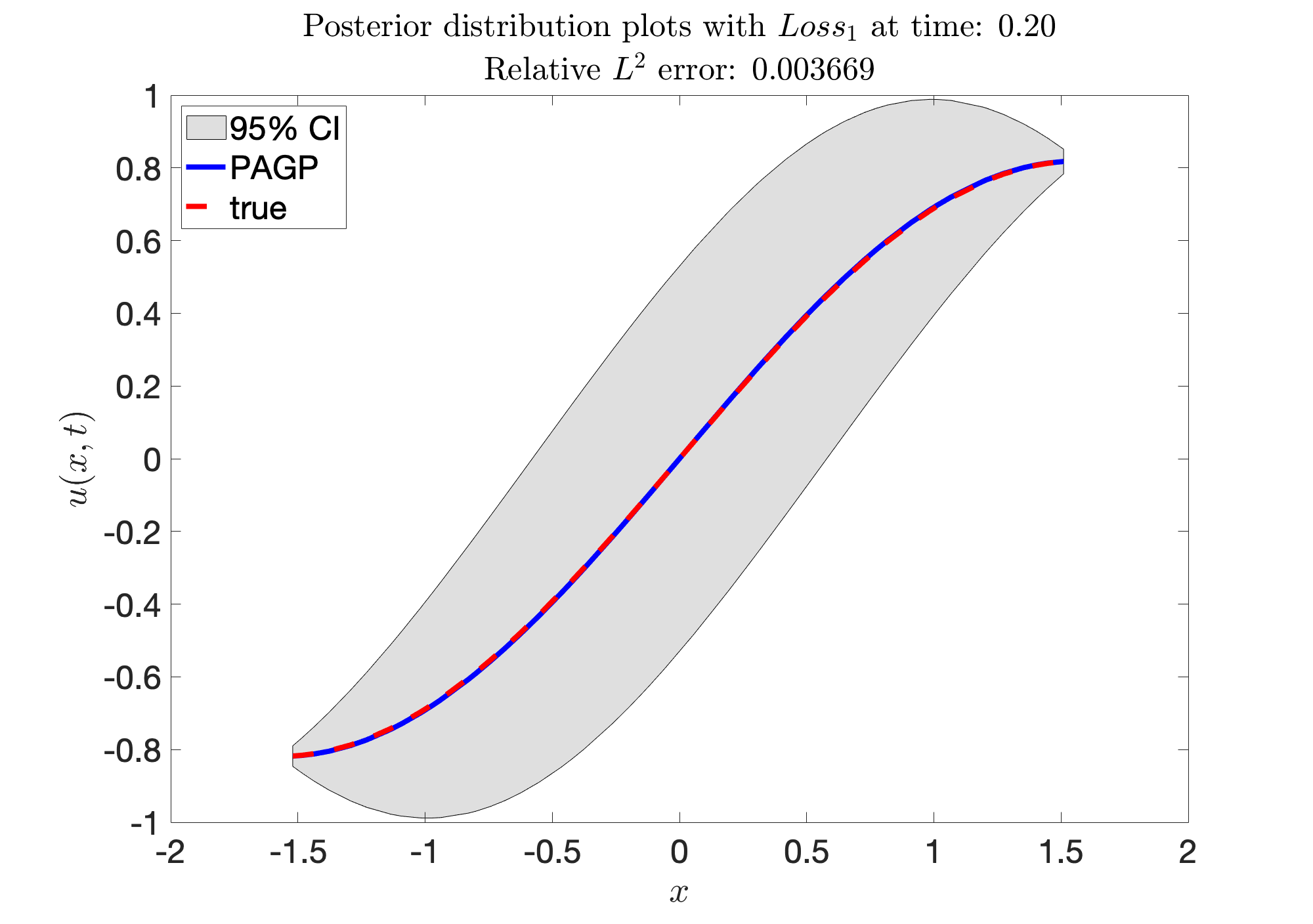}
\label{fig:Heat1d_predvd}}
\qquad 
\subfloat[c][]{
\includegraphics[width=0.45\textwidth]{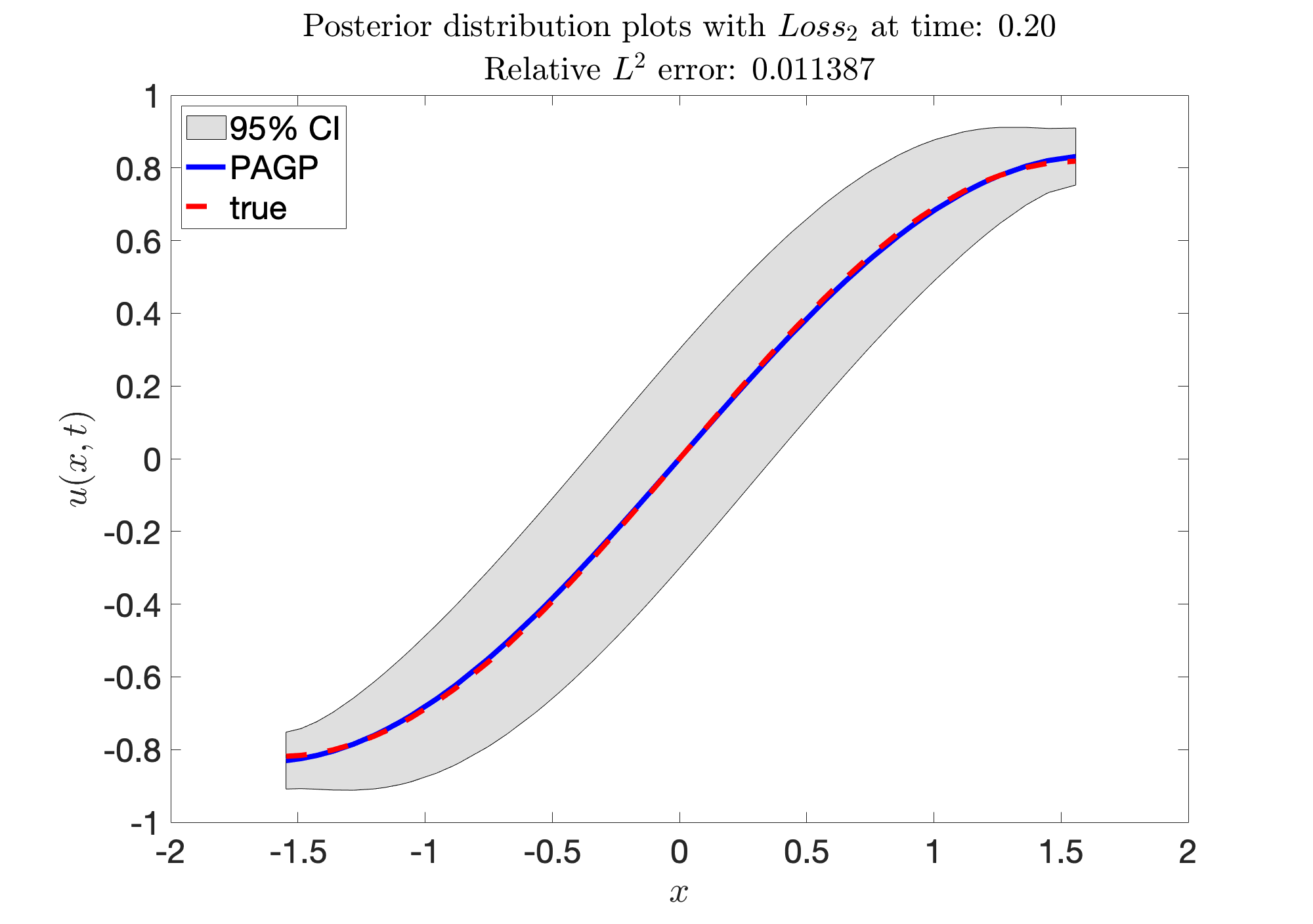}
\label{fig:Heat1d_predse}}
\qquad 
\subfloat[d][]{
\includegraphics[width=0.45\textwidth]{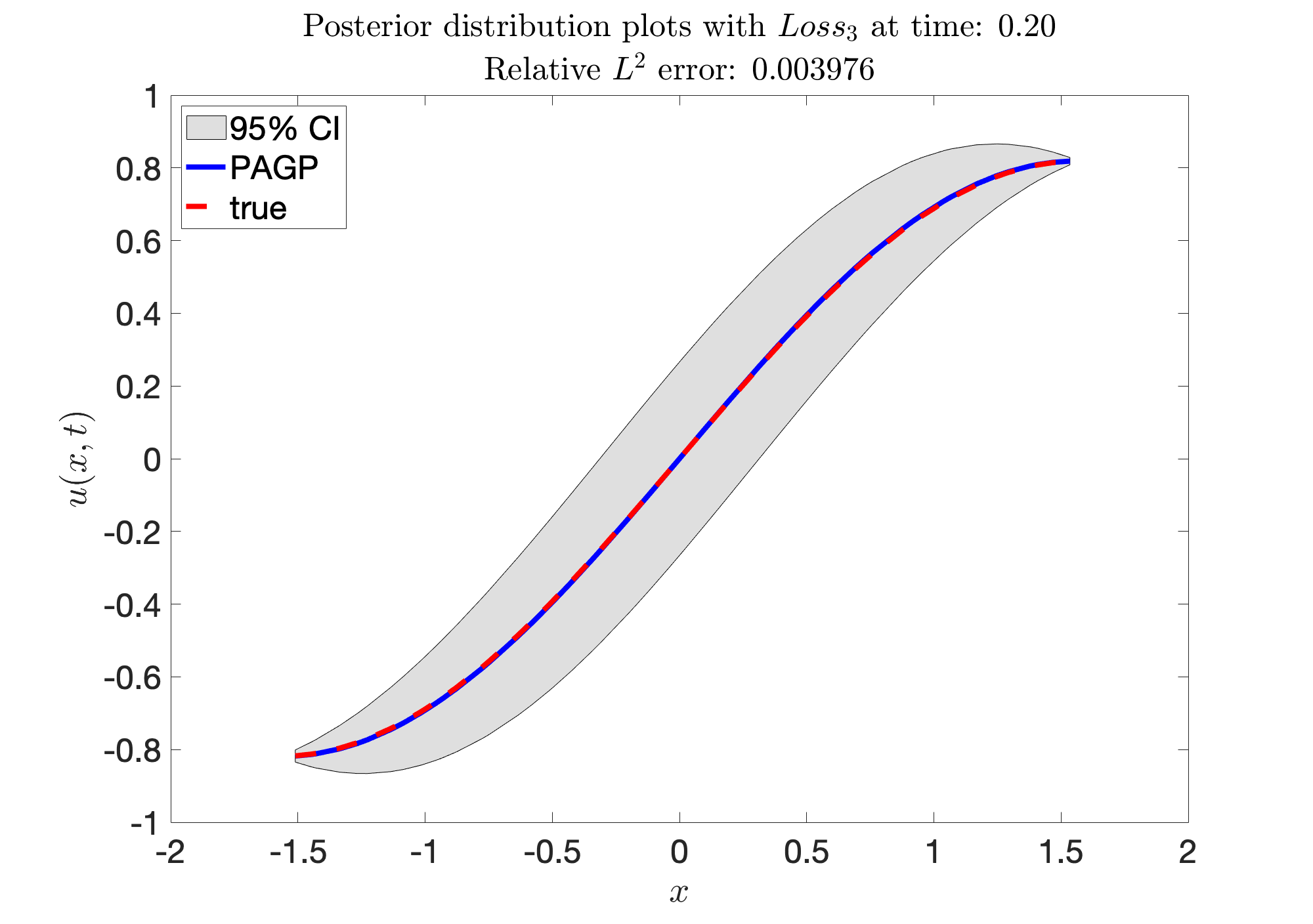}
\label{fig:Heat1d_predlog}}
\caption{Equation \ref{e20}'s initial condition plot along with posterior distribution plots of the solution at $t = 0.2$ using three loss functions. The number of training data is $60$, the number of collocation points is $900$ and the number of test points is $50$. The blue solid lines are the posterior distribution mean plots by PAGP continuous time model. The red dashed lines are the ground truth solution plots. The grey regions are the 95\% confidence intervals around the mean. (a): Initial condition and training samples; (b): plots generated by using Equation \ref{e13}; (c): plots generated by using Equation \ref{e14}; (d): plots generated by using Equation \ref{e15}.}
\label{fig:Pred_heat1}
\end{figure}

Figure \ref{fig:Pred_heat1} presents the initial data plots and posterior distribution plots of Equation \ref{e20} at times $t = 0.2$ with different loss functions. The loss function Equation \ref{e13} is used in part (b). The loss function Equation \ref{e14} is used in part (c) and Equation \ref{e15} is used in part (d). The red dashed line in the first figure is the plot of initial condition Equation \ref{e23}. Other red dashed lines are the ground truth generating solution plots at time $t = 0.2$. The blue solid lines represent the posterior distribution mean plots. And the shaded grey regions in each figure depict the 95\% confidence intervals around the posterior mean. We can see PAGP continuous time model with loss function $Loss_1$ can reconstruct the true solution more accurately because it achieves the smallest relative $L^2$ error. However, the 95\% confidence intervals is relatively big although there are 900 collocation points inside the domain. This indicates the model recover the solution with little confidence. Notice that the model is more confident for its predictions near boundaries since we have exact boundary conditions at $x=-\pi/2$ and $x=\pi/2$. There is no training data, i.e. information, inside the domain $[-\frac{\pi}{2}, \frac{\pi}{2}] \times [0,1]$, which is the main reason why the confidence intervals are so wide there. From part (d) of the figure, we can see the confidence intervals is smaller if loss function $Loss_3$ is utilized. This is because the marginal likelihood part of the loss function $Loss_3$ considers a trade-off between data-fit and model complexity. It doesn't favour the models that best fit the training data which results in a less posterior variance.

\begin{figure}[h]
\centering
\subfloat[a][]{
\includegraphics[width=0.45\textwidth]{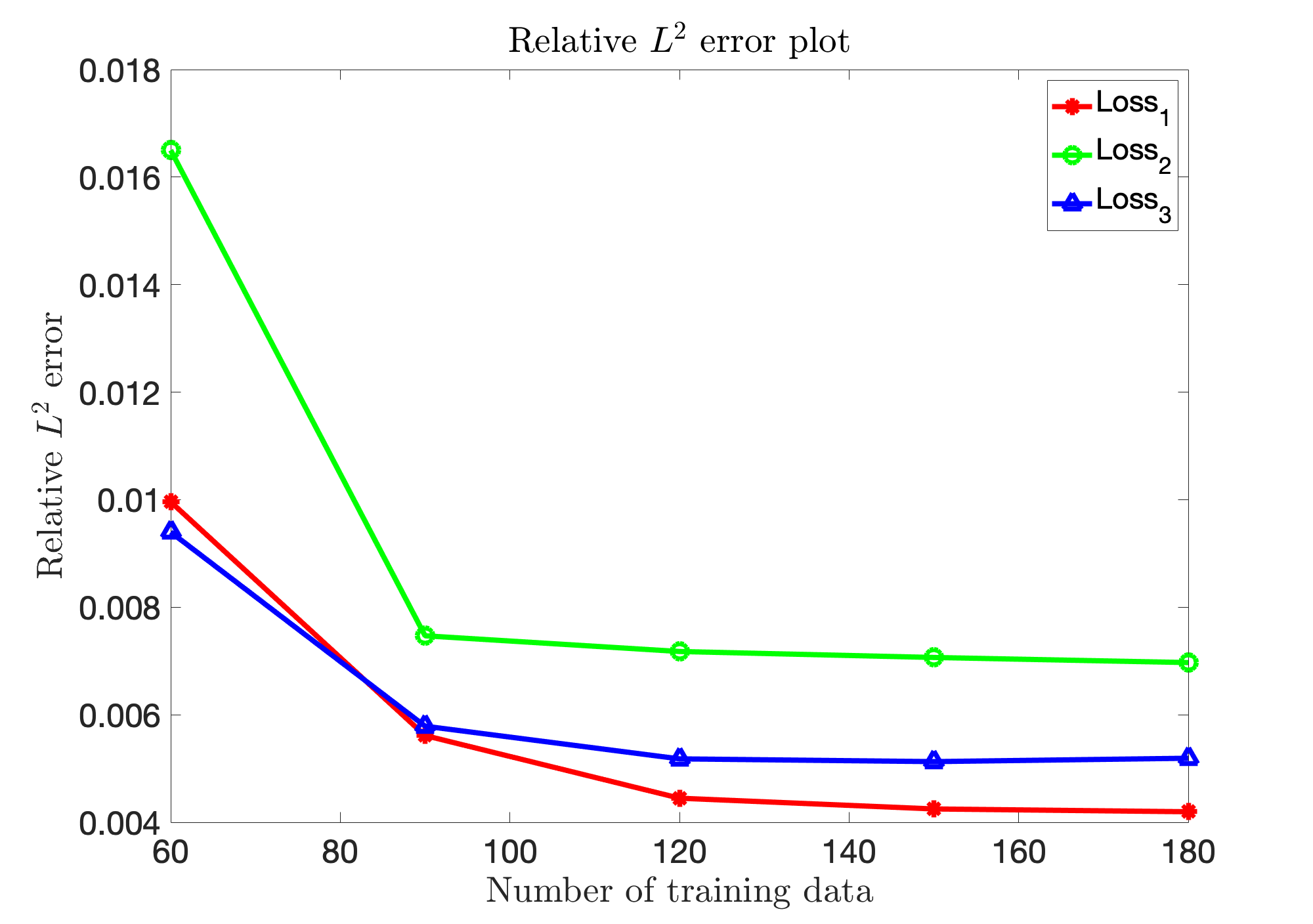}
\label{fig:Heat1d_l2t}}
\qquad 
\subfloat[b][]{
\includegraphics[width=0.45\textwidth]{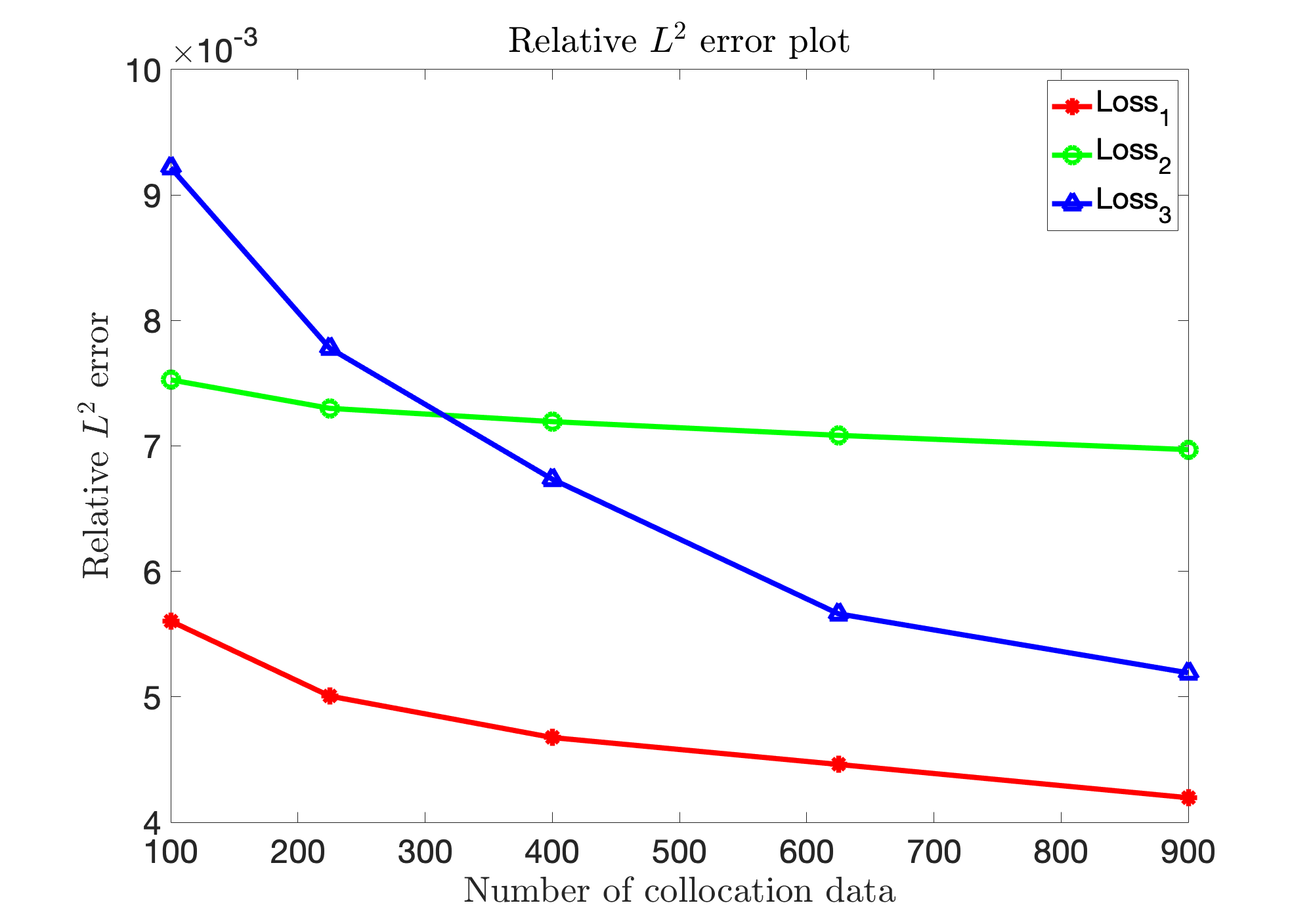}
\label{fig:Heat1d_l2c}}
\caption{Equation \ref{e20}'s relative $L^2$ error plots using different loss functions. The red lines are the plots of Equation \ref{e13}. The green lines are the plots of  Equation \ref{e14} and the blue lines are the plots of Equation \ref{e15}. (a): relative $L^2$ error plots with different number of training data. The number of collocation points is fixed to be 900; (b): relative $L^2$ error plots with different number of collocation points. The number of training data is fixed to be 180.}
\label{fig:L2_heat}
\end{figure}

Now, the model accuracy of the predictions with increasing number of training or collocation points is investigated. First we fix the number of collocation points and increase the number of training data. Figure \ref{fig:Heat1d_l2t} shows the relative $L^2$ error decreases as the number of training data increases no matter what loss functions is used. The number of collocation points is fixed to be 900. Moreover, the relative $L^2$ error with loss function $Loss_1$ is less than that with $Loss_2$ or $Loss_3$ at same number of training data. In Figure \ref{fig:Heat1d_l2c}, the number of training data is fixed to be 180. We can see the relative $L^2$ error decreases as the number of collocation data increase for all cases but with a different trend. Again, the red line shows that the relative $L^2$ error using $Loss_1$ is smaller than the other cases. These results tell us that increasing the number of training data or collocation points can both improve the model accuracy with different rates for all three loss functions. 

For the inverse problem, the boundary and initial conditions are given while finding coefficient $\lambda$ is our target. We first construct a training data set by generating $N=200$ points across the entire domain from exact solution with $\lambda = 1$. Note that there is no noise added to the training data. Then the continuous time model is trained using this data set by minimizing the loss function $Loss_1$, $Loss_2$ or $Loss_3$. The estimated coefficient $\lambda$ is $0.999904$ using loss function $Loss_1$, $1.002292$ using loss function $Loss_2$, and $0.999205$ using loss function $Loss_3$. So the errors are $0.0096\%$, $0.2292\% $ and $0.0795\%$ respectively. We can conclude that the continuous model can recover the unknown coefficient $\lambda$ in this example with all three loss functions. But the errors using loss function $Loss_1$ is smallest. Next, we perform the same experiment under the same setting. The training data are corrupted with 1\% uncorrelated Gaussian noises this time, the estimated coefficients are $1.001656$, $0.997973$ and $0.9955879$ using loss functions $Loss_1$, $Loss_2$ and $Loss_3$. So the errors are $0.1656\%$, $0.2027\%$ and $0.44121\%$ with loss function $Loss_1$, $Loss_2$ and $Loss_3$, respectively. See Table \ref{table:1} for the results. Note that the results are the means of $20$ repeated experiments under the same setting.

\begin{table}[ht!]
\centering
\begin{tabular}{||c c c||} 
 \hline
 & no noise  &  $1\%$ noise  
 \\ [0.5ex] 
  \hline
   True &  1     & 1   \\
 \hline
    Estimated coefficient  ($Loss_1$) & 0.999904  & 1.001656  \\  
 \hline
 Estimated coefficient  ($Loss_2$)    & 1.002292   & 0.997973 \\  
 \hline
 Estimated coefficient  ($Loss_3$)    & 0.999205   &   0.9955879 \\  
 \hline
\end{tabular}
\caption{Inverse problem for the one dimensional heat equation(Equation \ref{e20}): true coefficients and the estimated coefficients using loss functions $Loss_1$, $Loss_2$ and $Loss_3$ with no noise or 1\% Gaussian noises.}
\label{table:1}
\end{table}

\subsection{One dimensional burgers' equation}
In this section, we consider the following one dimensional Burger's equation which is a fundamental partial differential equation occurring in many areas of applied mathematics,

\begin{equation}\label{e25}
    \frac{\partial}{\partial t}u + u  \frac{\partial}{\partial x}u - \mu \frac{\partial^2}{\partial x^2}u = 0
\end{equation}
with boundary conditions,
\begin{equation}\label{e26}
     u(-1, t) = u(1, t) = 0
\end{equation}
where $t > 0$ and $\mu$ is the diffusion coefficient. The initial condition is set to be,
\begin{equation}\label{e27}
    u(x, 0) = - sin(\pi x)
\end{equation}

For the forward problem, we fix the diffusion coefficient $\mu = 0.01/\pi$. The second order Adams-Bashforth time discretization scheme is applied and the step size is set to be $0.01$s. We first sample $N_t = 50$ training data points at $t=0$ from the initial and boundary conditions. Then $N_c = 25$ collocation points are randomly sampled in spatial domain. The covariance function we use for this example is the so-called neural network function, 
\begin{equation}\label{e28}
    k(x, x';\theta) = \frac{2}{\pi} sin^{-1} \left( \frac{2(\sigma_0^2 + \sigma^2 x x')}{\sqrt{(1 + 2(\sigma_0^2 + \sigma^2 x^2))(1 + 2(\sigma_0^2 + \sigma^2 x'^2)}} \right)
\end{equation}
where $\theta = (\sigma_0, \sigma)$ is the set of hyper-parameters. The reason why this covariance function is more appropriate can be found in \cite{pang2019neural}.

\begin{figure}[h]
\centering
\includegraphics[width=1\textwidth]{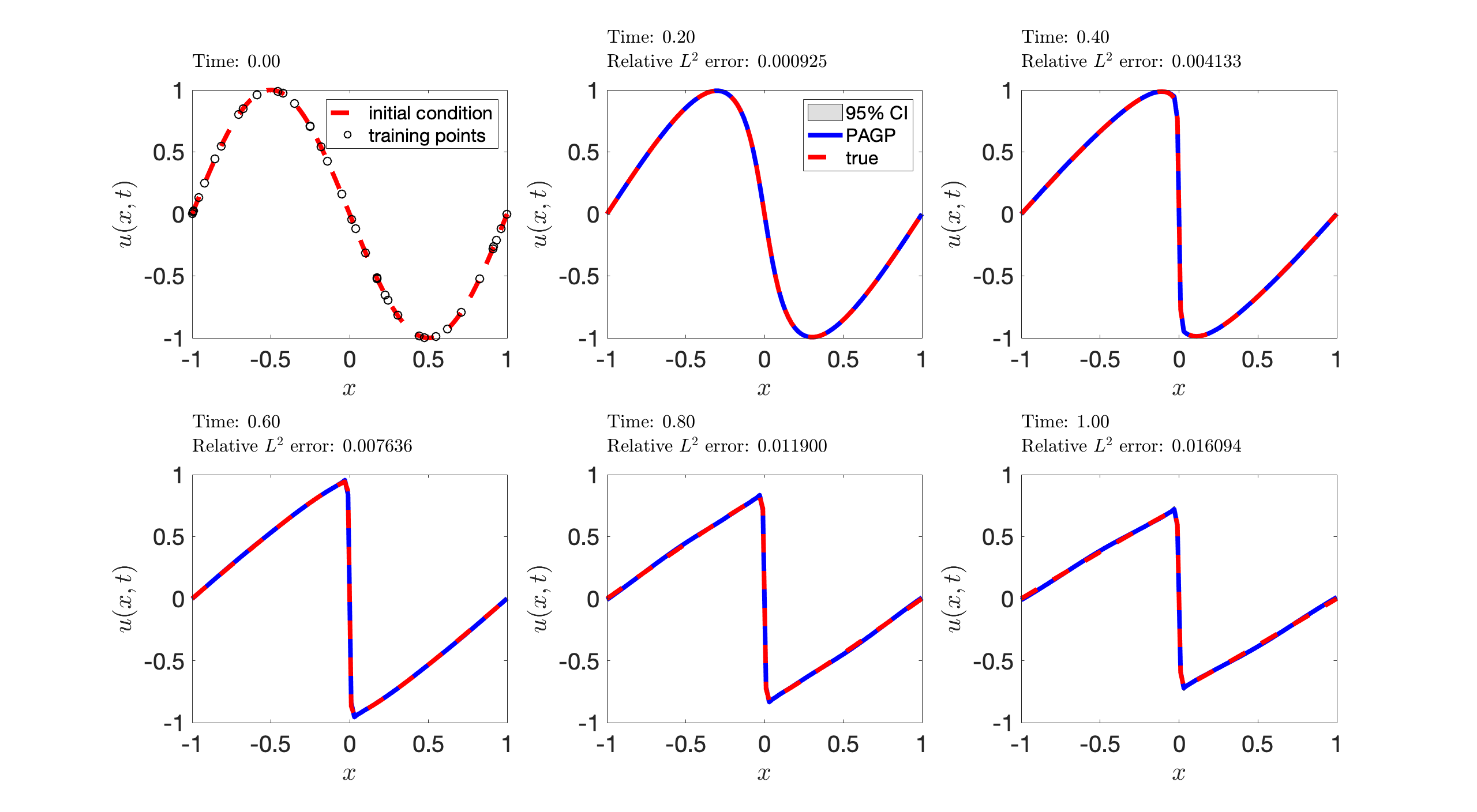}
\caption{Burger's equation's initial condition plot along with posterior distributions of the solution at five different times using loss function Equation \ref{e13}. The time discretization scheme is chosen to be second order Adams–Bashforth method. The number of training data is $50$ and the number of collocation data is $25$ at each time step. The blue solid lines are the prediction plots by PAGP discrete time model. The red dashed lines are the ground true prediction plots. The grey regions represent the 95\% confidence intervals.}
\label{fig:Pred_burger1}
\end{figure}

The loss function $Loss_1$, $Loss_2$ or $Loss_3$ is minimized using BFGS algorithm to find the optimal hyper-parameters. The initial weight coefficients are chosen to be $\omega = 1$ for all loss functions. The rate factors are $1.2$ for $Loss_1$ and $Loss_2$, and $0.6$ for $Loss_3$. The maximal iteration number for updating the weight coefficients is $5$. Here, we present the best prediction results which is obtained using loss function $Loss_1$. Figure \ref{fig:Pred_burger1} present the initial training data and posterior distributions at different time snapshots. The blue solid lines are posterior distribution means at each time and the red dashed lines are true solutions. The grey regions are 95\% confidence intervals, i.e. $\pm2$ standard deviations band around the mean. We can see from the plots that the posterior distributions fit the true solution well. The 95\% confidence intervals show the model is pretty confident of its predictions at all spatial locations. Note that the discontinuity developed over time by the small value of diffusion coefficient make the forward problem hard to solve for the classical numerical methods.

\begin{figure}[h]
\centering
\subfloat[a][]{
\includegraphics[width=0.45\textwidth]{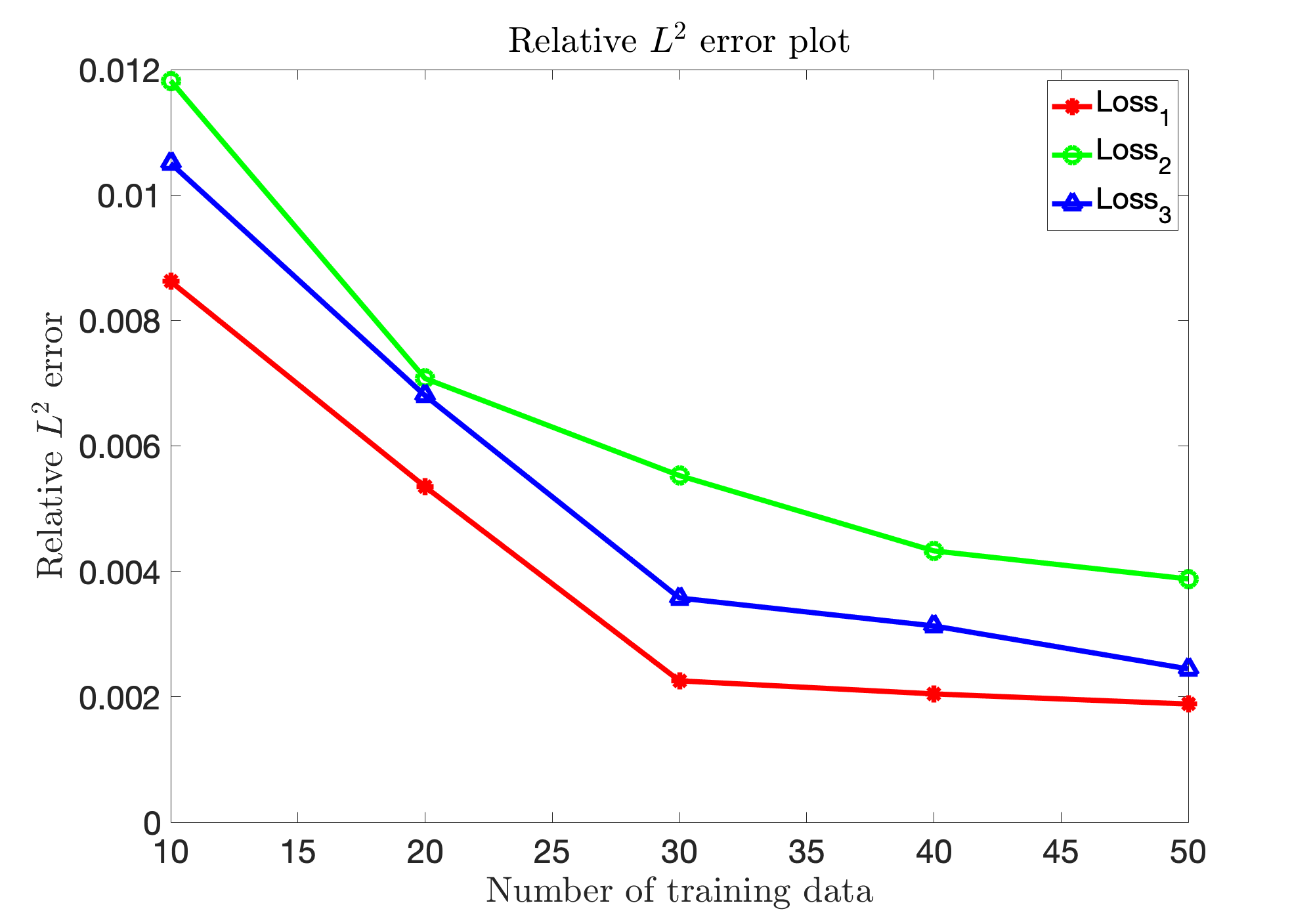}
\label{fig:Burger1d_l2t}}
\qquad 
\subfloat[b][]{
\includegraphics[width=0.45\textwidth]{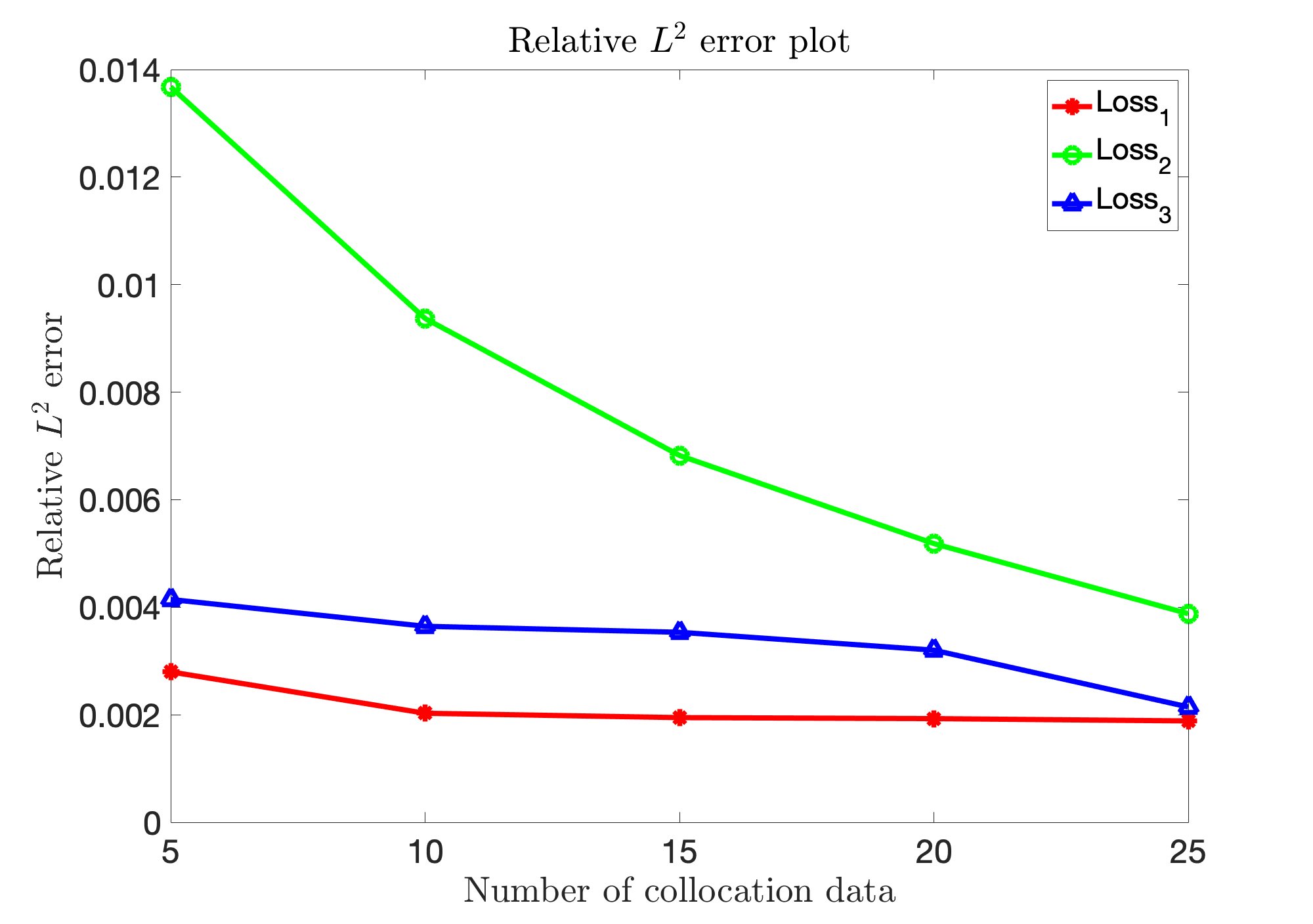}
\label{fig:Burger1d_l2c}}
\caption{Burger's equation relative $L^2$ error plots using different loss functions. The red lines are the plots of Equation \ref{e13}. The green lines are the plots of  Equation \ref{e14} and the blue lines are the plots of Equation \ref{e15}. (a): relative $L^2$ error plots with different number of training data. The number of collocation points is fixed to be 25; (b): relative $L^2$ error plots with different number of collocation points. The number of training points is fixed to be 50.}
\label{fig:L2_burger1}
\end{figure}

Figure \ref{fig:L2_burger1} illustrates the model accuracy with increasing number of training or collocation points using loss function $Loss_1$, $Loss_2$ and $Loss_3$. Part (a) shows the relative $L^2$ error is decreasing with the increasing number of training data for all three loss functions. The number of collocation points is fixed to be 25. We can also see that results using loss function $Loss_1$ achieve the smallest relative $L^2$ error if the number of training data is the same. Part (b) presents the relative $L^2$ error is decreasing with the increasing number of collocation points. The number of training points is fixed to be 50. The relative $L^2$ error using $Loss_1$ is the smallest. We can see a similar trend for the plots using $Loss_1$ and $Loss_3$, while the relative $L^2$ error using $Loss_2$ is bigger when the number of collocation points is small. Also, the improvement to the relative $L^2$ error by increasing the number of collocation points for $Loss_1$ and $Loss_3$ is less than the impact of increasing the number of training points.

\begin{figure}[h]
\centering
\subfloat[a][]{
\includegraphics[width=0.45\textwidth]{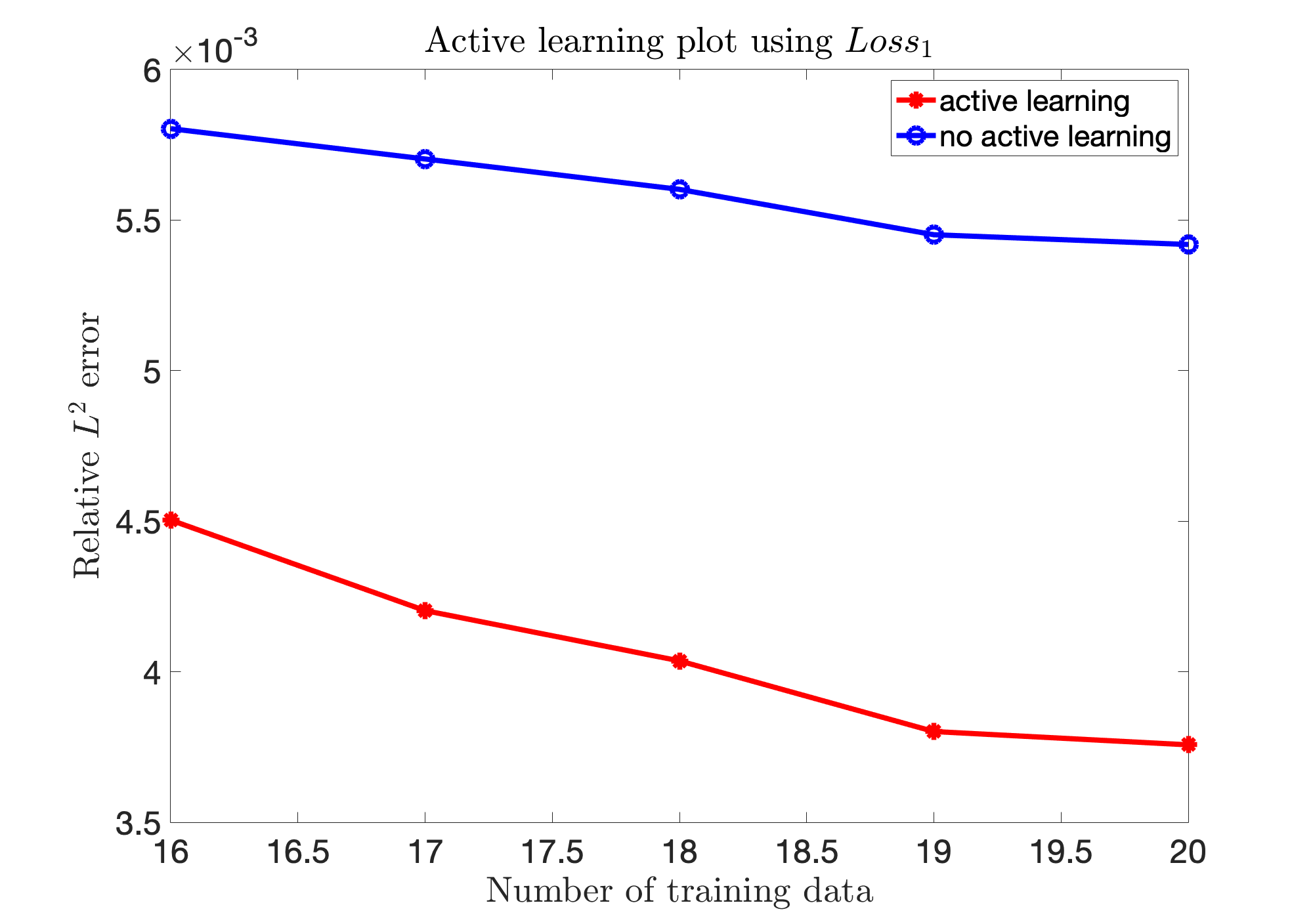}
\label{fig:ac_burgervd}}
\qquad 
\subfloat[b][]{
\includegraphics[width=0.45\textwidth]{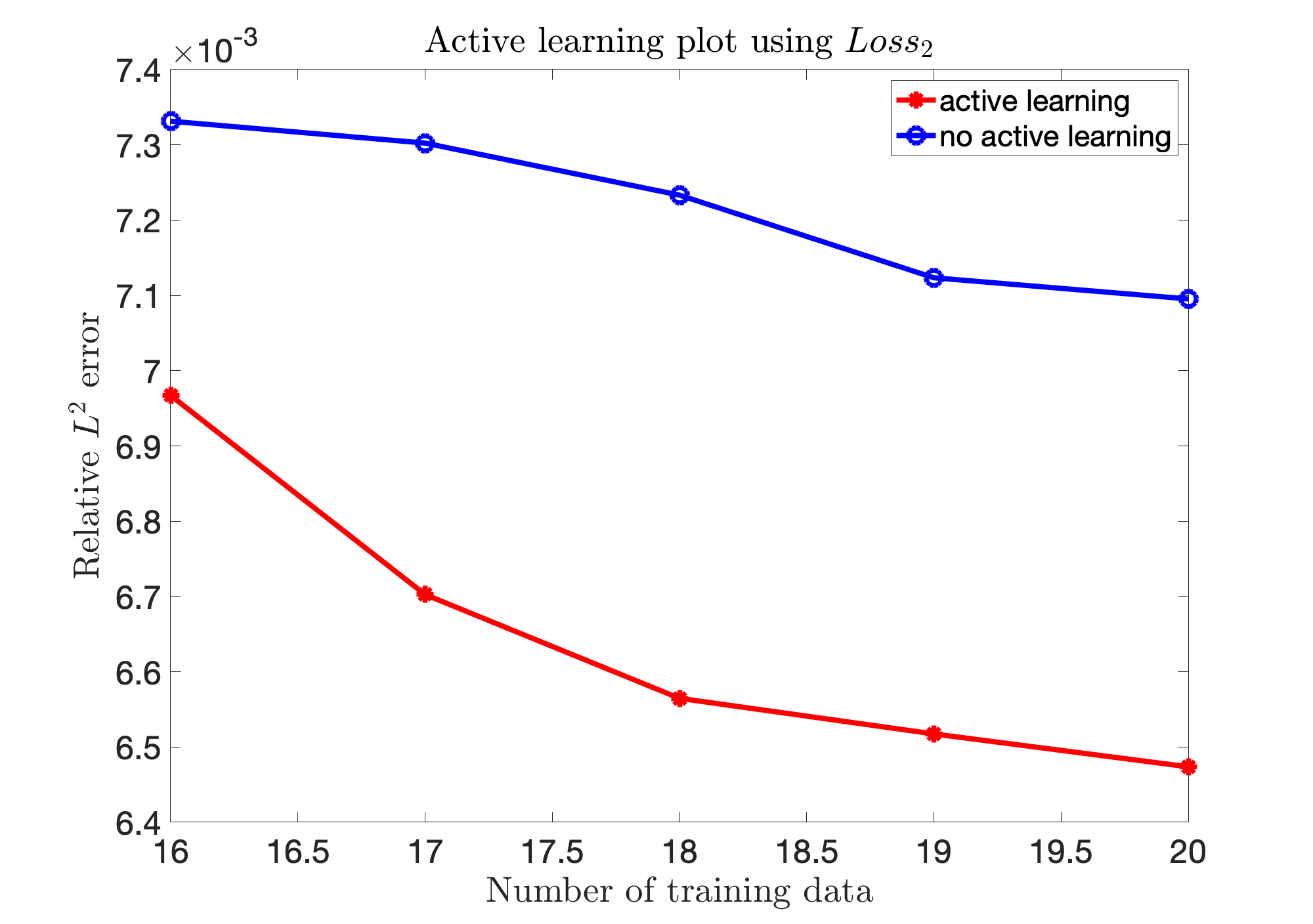}
\label{fig:ac_burgerse}}
\qquad 
\subfloat[c][]{
\includegraphics[width=0.45\textwidth]{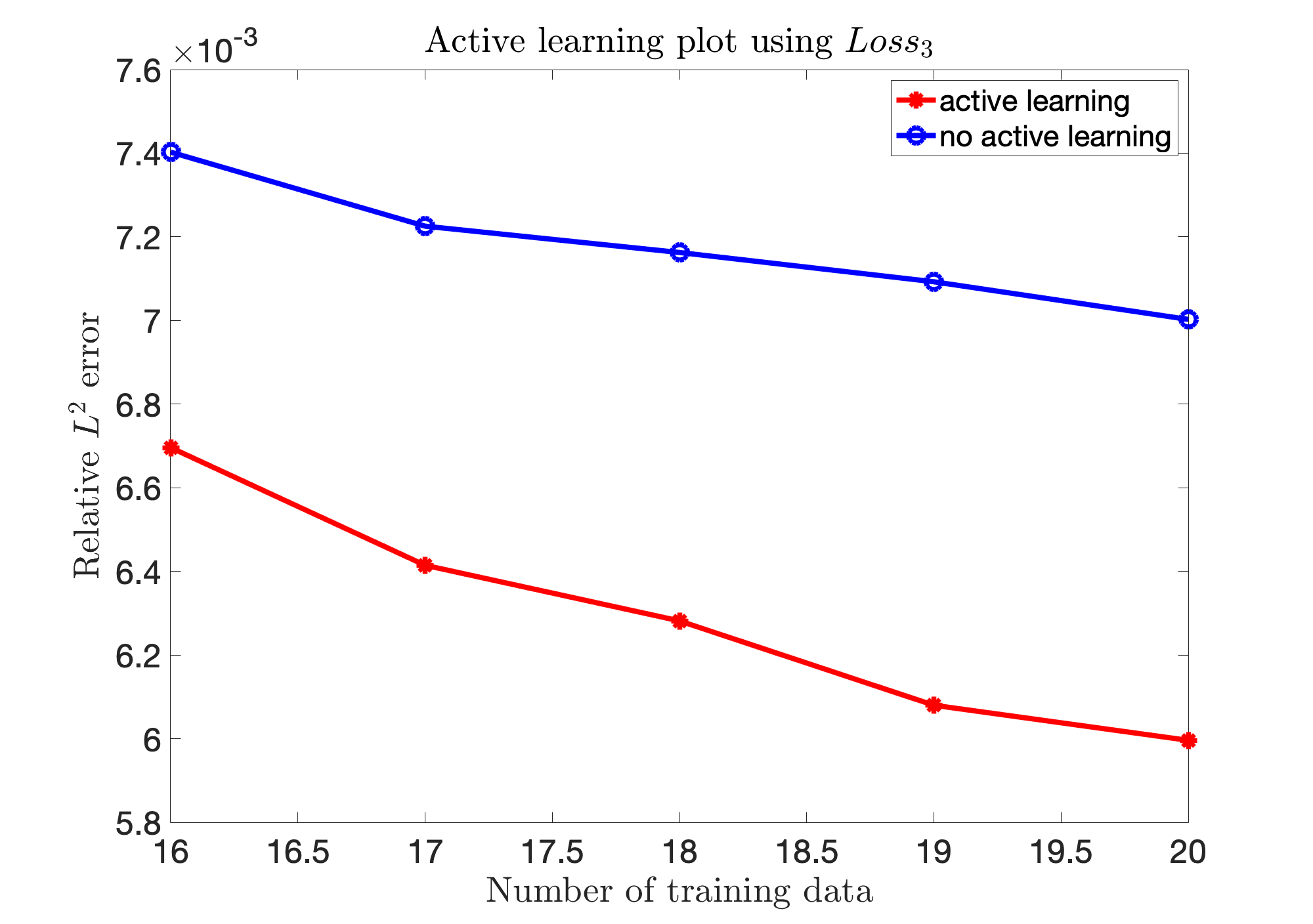}
\label{fig:ac_burgerlog}}
\caption{Relative $L^2$ error plots of Burger's equation's solutions using different number of training data via active learning. The blue solid lines are the relative $L^2$ error plots with active learning and the red solid lines are the relative $L^2$ error plots without active learning. (a): loss function used is Equation \ref{e13}; (b): loss function used is Equation \ref{e14}; (c): loss function used is Equation \ref{e15};}
\label{fig:ac_burger}
\end{figure}

As we mentioned in Section \ref{AC}, the active learning scheme can be applied in the discrete time model. We start from 15 training points and add one additional point per iteration by our active learning scheme for a total of five iterations. The criterion for choosing the next point is the location where it achieves the maximal variance. For a comparison purpose, the same experiment is performed but the additional point are randomly drawn in each iteration. Figure \ref{fig:ac_burger} shows the results. Part (a) is the results when loss function $Loss_1$ is applied. The blue solid line is the relative $L^2$ error plot with active learning and the red dashed line is relative $L^2$ error plot without active learning. The relative $L^2$ error is smaller and decreases faster using active learning scheme. Similar results can be found in part (b) and (c) when loss functions $Loss_2$ and $Loss_3$ are applied respectively. We can conclude that model performance is indeed improved when our active learning scheme is applied.

For the inverse problem, we are given the boundary and initial conditions. The target is to find the coefficient $\mu$ in the Burger's equation (Equation \ref{e25}). Following the method introduced in Section \ref{s242}, we first set the two time step $t_1 = 0.1$ and $t_2 = 0.3$ at where the observations are available. Then two sets of training data set at $t_1$ and $t_2$ are generated from the exact solutions with $\mu = 0.3$. The number of training data in above two training sets is set to be $N_1 = N_2 = 20$. Then the proposed model is trained using these two data sets by minimizing the loss function $E(\mu)$, i.e. Equation \ref{e17}. The estimated coefficients $\mu$ are $0.301233$ if there is no noise involved in the training data. The relative errors are $0.1233\%$. When there is $1\%$ Gaussian noises, the estimated coefficients $\mu$ are $0.29132$. The errors are $0.868\%$. We can conclude that the discrete time model can recover the unknown coefficient $\mu$ in the inverse problem setting with relatively small errors even there are $1\%$ Gaussian noises in the training data. See Table \ref{table:2} for the results. Note that the results are the means of $20$ repeated experiments under the same setting.

\begin{table}[ht!]
\centering
\begin{tabular}{||c c c||} 
 \hline
 & True coefficient      &   Estimated coefficient
 \\ [0.5ex] 
  \hline
  no noise &  0.3     &   0.301233 \\
 \hline
$1\%$ noise    & 0.3   & 0.29132 \\  
 \hline
\end{tabular}
\caption{Inverse problem of the Burger's equation (Equation \ref{e25}): true coefficient and the estimated coefficient with no noise or $1\%$ Gaussian noises.}
\label{table:2}
\end{table}


\subsection{Two dimensional heat equation}
For the hybrid model, let's first consider the following two dimensional heat equation,
\begin{equation}\label{e29}
    \frac{\partial}{\partial t}u = \frac{\partial^2}{\partial x^2}u + \frac{\partial^2}{\partial y^2}u,
    \hspace{3mm} x,y \in [0,\pi], \hspace{3mm} t \in [0, 0.5]
\end{equation}
with initial condition,
\begin{equation}\label{e30}
    u(0, x, y) = sin(x) sin(y)
\end{equation}
and boundary conditions,
\begin{equation}\label{e31}
    u(t, 0, y) = u(t, \pi, y) = 0, \hspace{3mm}
    u(t, x, 0) = u(t, x, \pi) = 0
\end{equation}
The true solution for this heat equation is,
\begin{equation}\label{e32}
    u(t, x, y) = e^{-2t} sin(x) sin(y)
\end{equation}

\begin{figure}[h]
\centering
\includegraphics[width=1\textwidth]{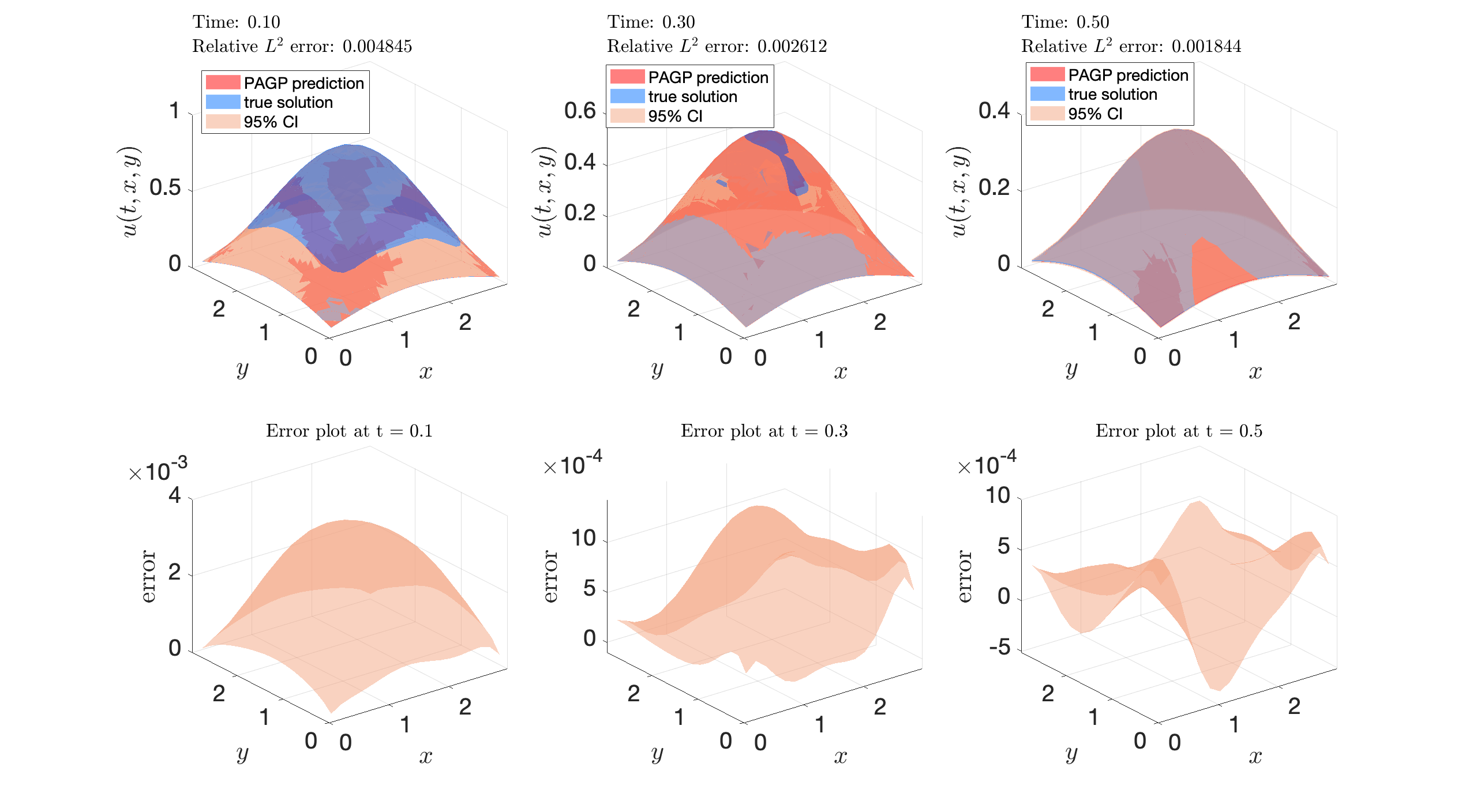}
\caption{Equation \ref{e29}'s posterior distribution along with corresponding error plots at three different times $t = [0.1, 0.3, 0.5]$ using loss function Equation \ref{e13}. The time discretization scheme in the discrete step is the second order Adams–Bashforth method. The number of training data is $30$ and the number of collocation data is $400$ at each time step. In continuous step, the number of samples from initial and boundary conditions is $425$ and the number of collocation points is $1000$. The number of final test points is $3240$. For the top three plots, the red surfaces are the posterior mean plots by PAGP hybrid model. The blue surfaces are the ground true prediction plots and the two orange surfaces represent plus/minus two standard deviations band around the posterior mean. The bottom three plots are the difference between the true solution and PAGP prediction at test points at corresponding times.}
\label{fig:Pred_heat2d}
\end{figure}

For this example, we only consider the forward problem, i.e. finding the solution for Equation \ref{e29}. As introduced in Section \ref{Hybrid}, the hybrid model is a two-step process. In the first discrete step of the hybrid model, we apply the second order Adams-Bashforth method with time step $\Delta t = 0.05$s. So the number of coarse layers is $10$. In each layer, we use $25$ evenly spaced grid points as the test points for this step and this set of test points is exactly part of the training set in the second step. The number of training points is $N_t = 30$ and the number of collocation points is $N_c = 400$. The predictions at above test points can be obtained by applying the discrete time model. Those points together with $25$ randomly drawn samples from initial condition (Equation \ref{e30}) and $400$ randomly drawn samples from boundary conditions (Equation \ref{e31}) form the training set for the continuous step. The collocation points is set to be $1000$ for this step. The number of test points is $3240$ grid points in the three dimensional space $\Omega = (t, x, y)$ because the temporal dimension is viewed as the same as the two spatial dimensions in this step where the continuous time model is applied. Here we conduct the experiments with all three loss functions $Loss_1$, $Loss_2$ and $Loss_3$. Their weight coefficients are adjusted similarly as in previous numerical examples. The best performance under the measurement of relative $L^2$ error is the one using $Loss_1$. The posterior distribution plots at $t = [0.1, 0.3, 0.5]$s and their corresponding error plots is shown in Figure \ref{fig:Pred_heat2d}. For the top figures, the red surfaces are the posterior distribution mean plots by PAGP hybrid model at different times. The blue surfaces are the ground true prediction plots. The two orange surfaces represent plus/minus two standard deviations band around the posterior means. The bottom figures are the plots of the difference between true predictions and PAGP predictions at the test points. We can see from the figures the PAGP hybrid model finds the solution with small relative $L^2$ errors and also achieves small variances.

\begin{figure}[h]
\centering
\subfloat[a][]{
\includegraphics[width=0.45\textwidth]{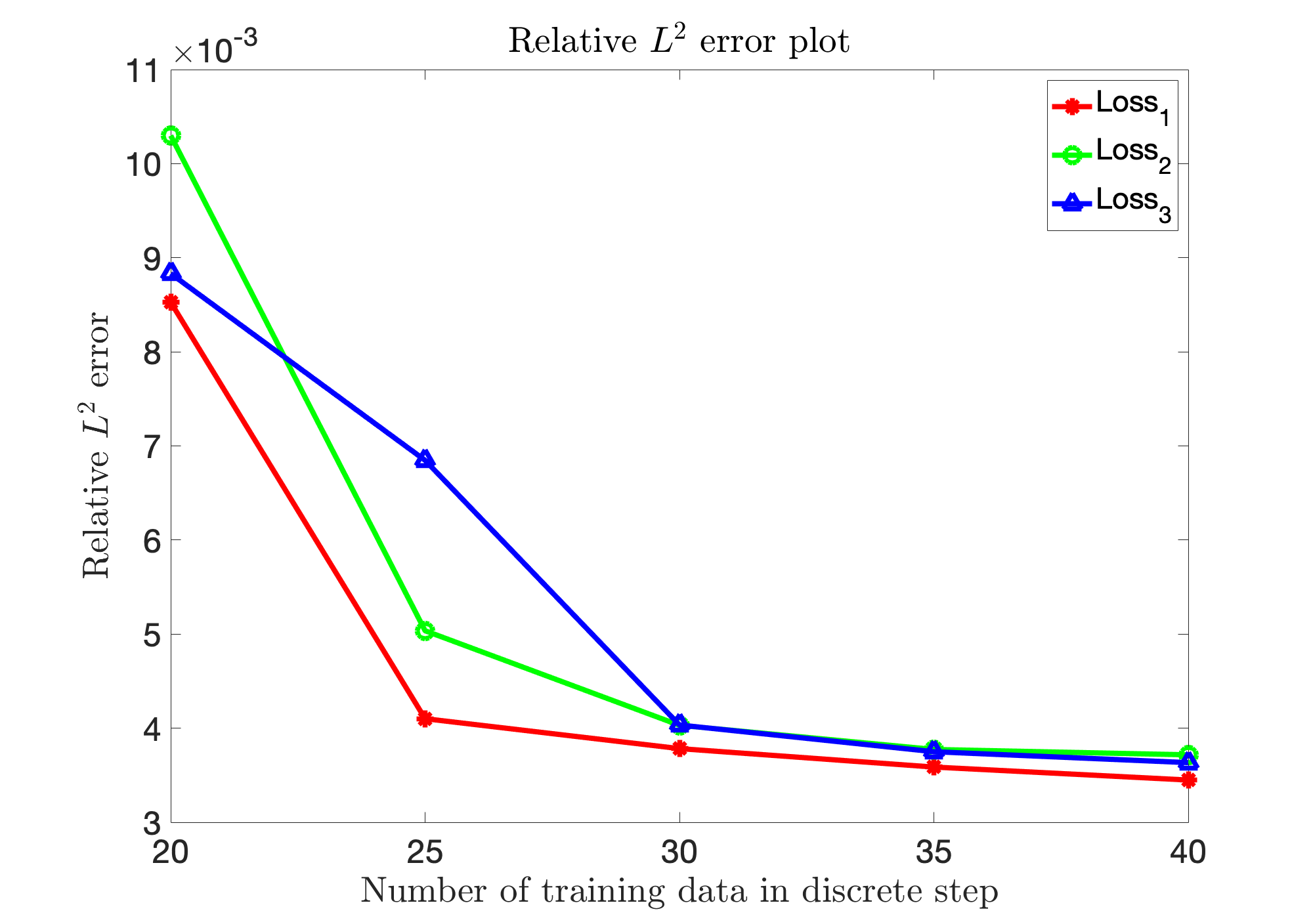}
\label{fig:Heat2d_l2t}}
\qquad 
\subfloat[b][]{
\includegraphics[width=0.45\textwidth]{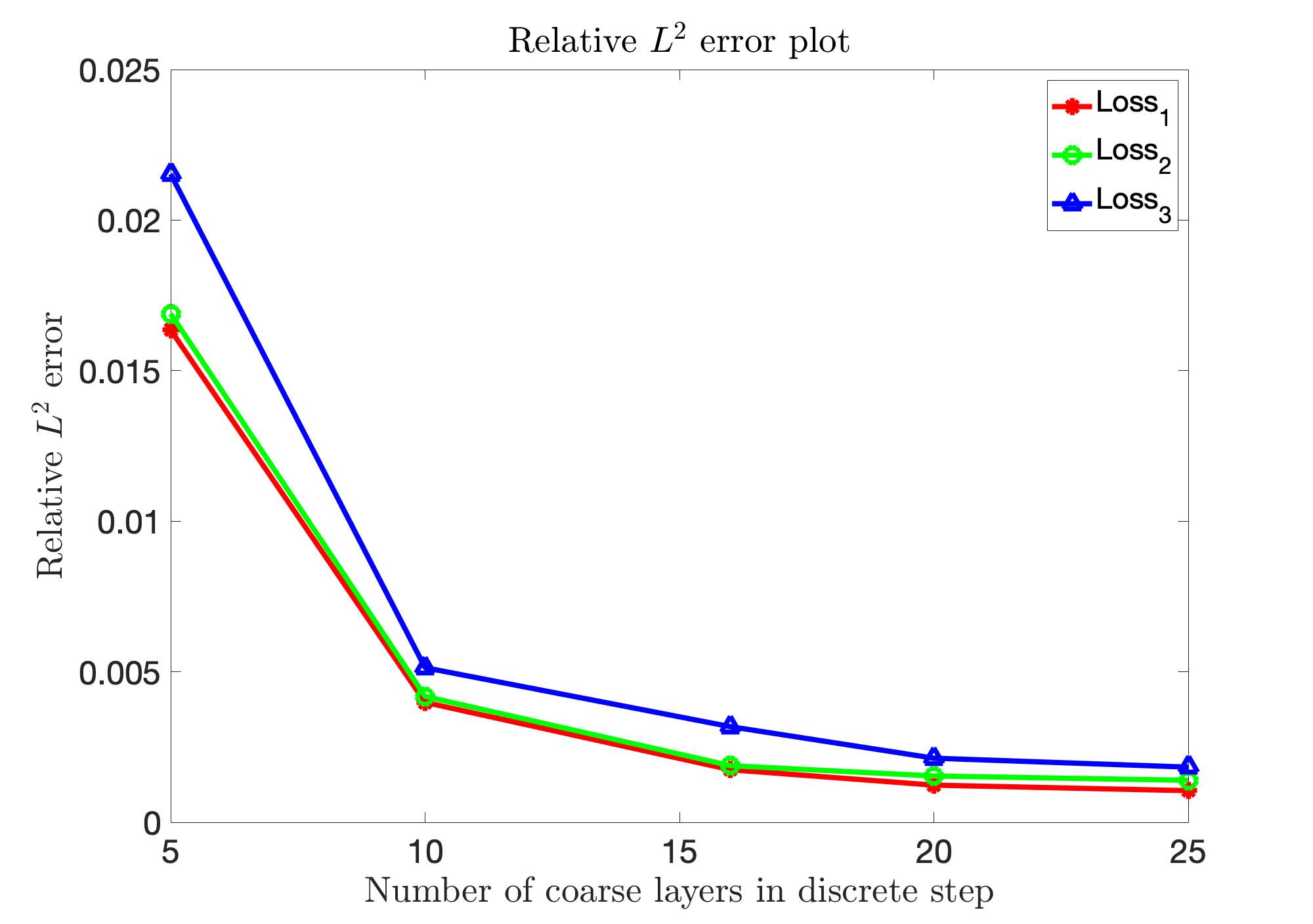}
\label{fig:Heat2d_l2c}}
\caption{Equation \ref{e29}'s relative $L^2$ error plots using different loss functions. The red lines are the plots of Equation \ref{e13}. The green lines are the plots of  Equation \ref{e14} and the blue lines are the plots of Equation \ref{e15}. (a): relative $L^2$ error v.s. the number of training data in the discrete step. The number of coarse layers in discrete step is fixed to be $10$; (b): relative $L^2$ error v.s. the number of coarse layers in the discrete step. The number of training points is fixed to be 30. For (a) and (b), the number of collocation points is $400$ in the discrete step and $1000$ in the continuous step. $3240$ evenly spaced grid test points in the whole domain is used to test the model performance.}
\label{fig:Heat2dL2}
\end{figure}

Next, we investigate the relationship between prediction accuracy and the number of training data or the number of coarse layers in the discrete step of hybrid model. The measure used for this purpose is the relative $L^2$ error of the predictive solutions at the test points. Figure \ref{fig:Heat2dL2} presents the results using loss functions $Loss_1$, $Loss_2$ and $Loss_3$. Part (a) shows the relative $L^2$ error is decreasing with the increasing number of training data for all loss functions in the first step. The number of coarse layers in discrete step is $10$. The number of collocation points is $400$ in the discrete step and $1000$ in the continuous step. Also, the relative $L^2$ error using $Loss_1$ is the smallest if the number of training data is the same. Part (b) presents the relative $L^2$ error is decreasing with the increasing number of coarse layers for three loss functions. The number of training points is fixed to be $30$. Again, we can see that the relative $L^2$ error wiht $Loss_1$ is smallest among all three loss functions.


\subsection{One dimensional advection equation}
In this example, we consider the advection equation in one space dimension,
\begin{equation}\label{e33}
    \frac{\partial}{\partial t}u + \frac{\partial}{\partial x}u =  0,
    \hspace{3mm} x \in [0,1], \hspace{3mm} t \in [0, 0.5]
\end{equation}
with initial condition,
\begin{equation}\label{e34}
    u(0, x) = 2 sin(\pi x)
\end{equation}
and boundary conditions,
\begin{equation}\label{e35}
    u(t, 0) = -2 sin(\pi t), \hspace{3mm}
    u(t, 1) = 2 sin(\pi t)
\end{equation}
The true solution for this advection equation is,
\begin{equation}\label{e36}
    u(t, x) = 2 sin(\pi (x - t))
\end{equation}

\begin{figure}[h]
\centering
\includegraphics[width=1\textwidth]{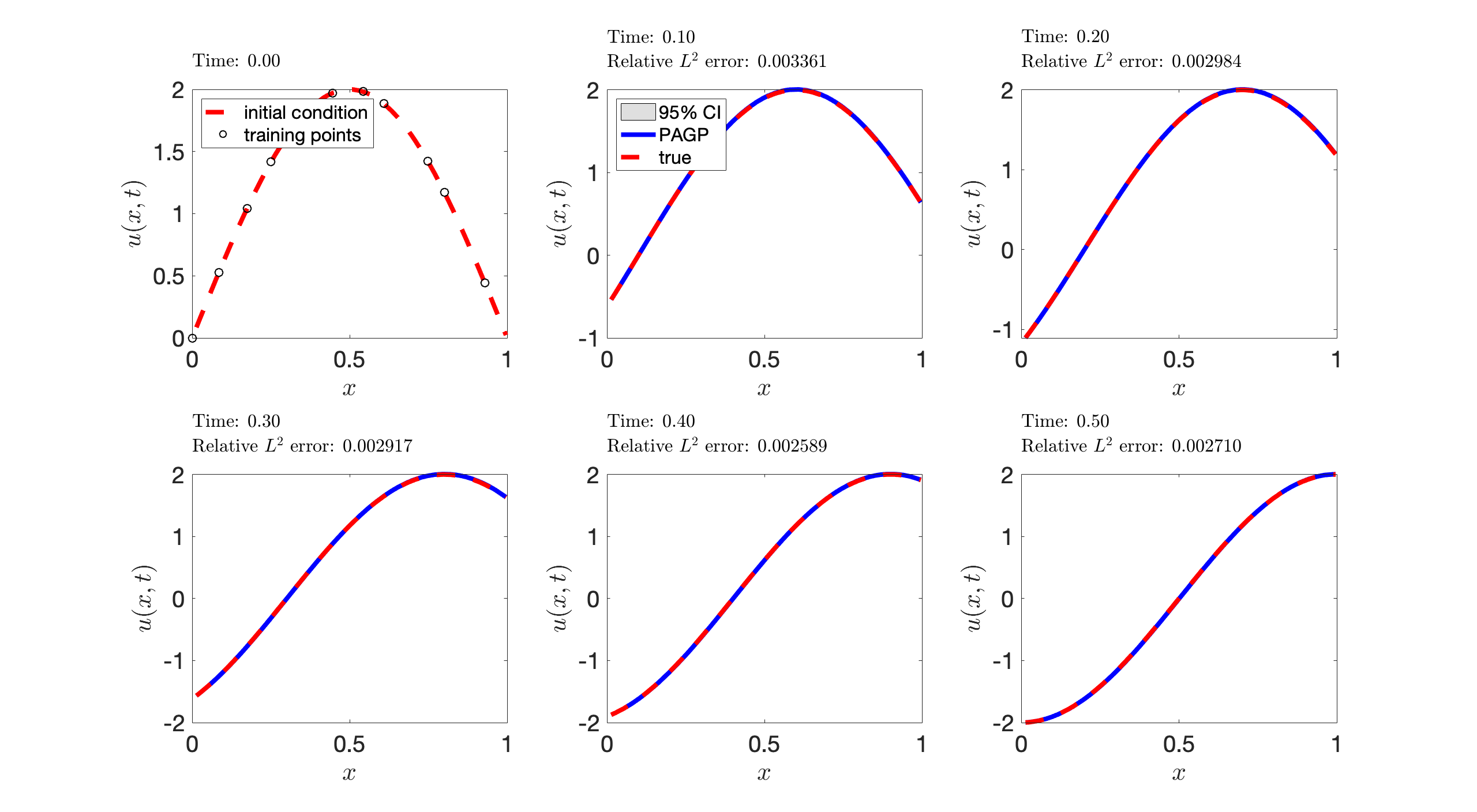}
\caption{Advection equation's initial condition plot along with posterior distributions of the solution at five different times using loss function Equation \ref{e13}. The time discretization scheme in the discrete step is the second order Adams–Bashforth method. The number of training data is $25$ and the number of collocation data is $30$ at each time step. In continuous step, the number of samples from initial and boundary conditions is $50$ and the number of collocation points is $900$. The blue solid lines are the posterior distribution plots by PAGP hybrid model. The red dashed lines are the ground truth plots. The grey regions represent the 95\% confidence intervals.}
\label{fig:Pred_adv1}
\end{figure}

As in the previous experiment, we only consider the forward problem, i.e. finding the solution for Equation \ref{e33} using hybrid model. In first discrete step, we apply the second order Adams-Bashforth method with time step $\Delta t = 0.025$s. So the number of coarse layers is $20$. In each layer, $10$ evenly spaced points are used as the test points at each time step. Note that this set of points is used as part of the training set in second step. The number of training points is $N_t = 25$ and the number of collocation points is $N_c = 30$. Following the discrete time model, the prediction at above test points can be obtained. Those points together with $10$ randomly drawn samples from initial condition (Equation \ref{e34}) and $40$ boundary conditions (Equation \ref{e35}) form the training set for the continuous step. The collocation points is set to be $900$ for this step. Three loss functions $Loss_1$, $Loss_2$ and $Loss_3$ can be used and their weight coefficients are adjusted similarly as before. We show the posterior distribution plots at $t = [0.1, 0.2, 0.3, 0.4, 0.5]$s using loss function $Loss_1$ which achieves the best performance in Figure \ref{fig:Pred_adv1}. The first figure is the plot of initial condition. The blue solid lines in other figures are the PAGP posterior distribution plots at different times. The red dashed lines are the true solution plots. The grey regions are the 95\% confidence intervals.

\begin{figure}[h]
\centering
\subfloat[a][]{
\includegraphics[width=0.45\textwidth]{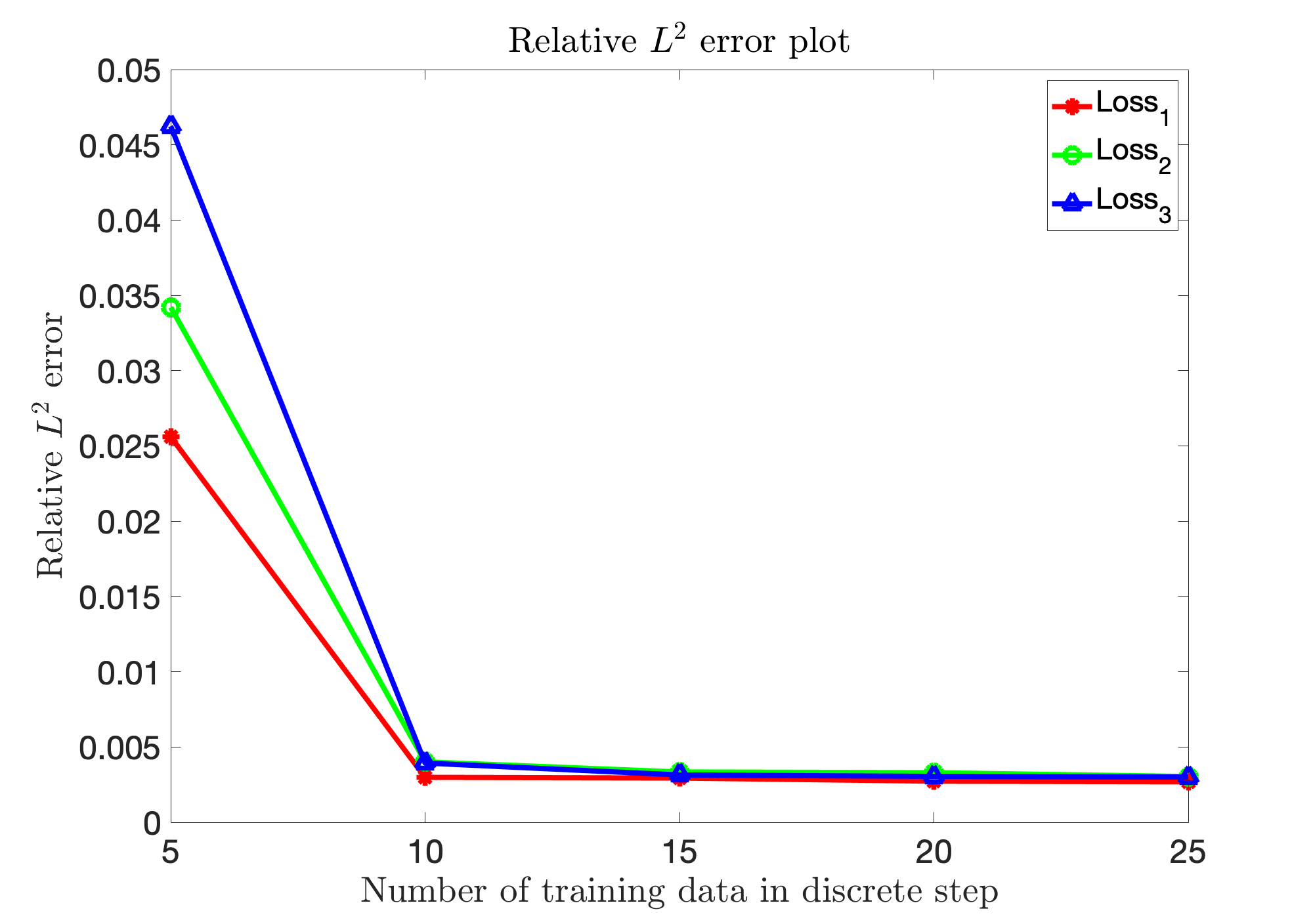}
\label{fig:ADV1d_l2t}}
\qquad 
\subfloat[b][]{
\includegraphics[width=0.45\textwidth]{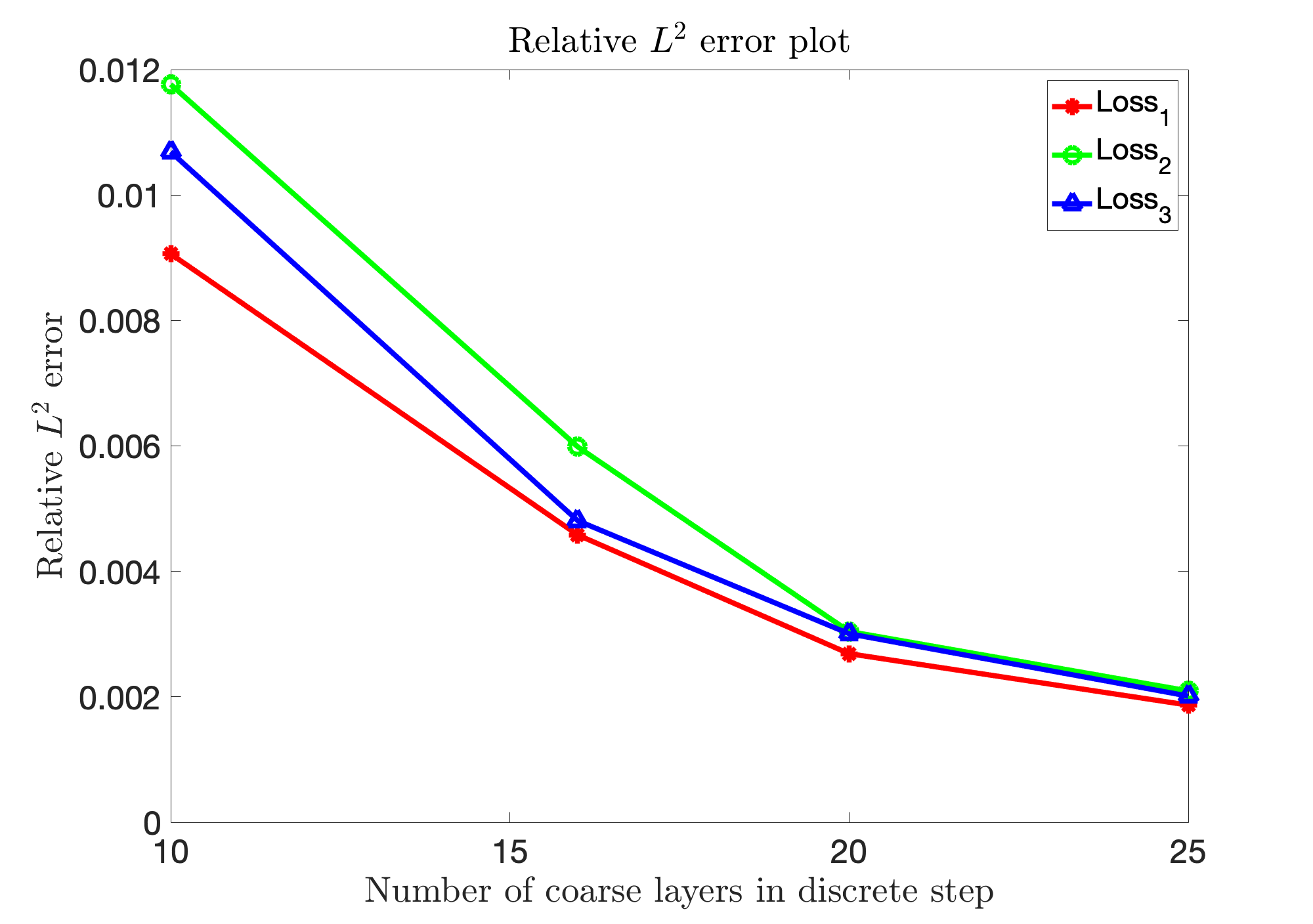}
\label{fig:ADV1d_l2c}}
\caption{Advection equation's relative $L^2$ error plots using different loss functions. The number of collocation points is $30$ in the discrete step and $900$ in the continuous step. $300$ test points are randomly chosen in the whole domain $[0,0.5] \times [0,1]$. The red lines are the plots of Equation \ref{e13}. The green lines are the plots of  Equation \ref{e14} and the blue lines are the plots of Equation \ref{e15}. (a): relative $L^2$ error plots with different number of training data in discrete step. The number of coarse layers in discrete step is fixed to be $20$; (b): relative $L^2$ error plots with different number of coarse layers in discrete step. The number of training data in discrete step is fixed to be $25$.}
\label{fig:Adv1dl2}
\end{figure}

Next, we investigate the relationship between prediction accuracy and the number of training data $n_t$ or the number of coarse layers $n_l$ in the discrete part of hybrid model. Two measures are introduced for this purpose: relative $L^2$ error of the predictive solutions and mean absolute PDE residue errors at test points. In the experiments, the number of collocation points is $30$ in the discrete step and $900$ in the continuous step. We randomly sample $300$ test points in the whole domain $[0,0.5] \times [0,1]$ to test the model performance. Figure \ref{fig:ADV1d_l2c} presents the results with the relative $L^2$ error. The red lines are the plots using $Loss_1$. The green lines are the plots using $Loss_2$ and the blue lines are the plots using $Loss_3$. For part (a), the time step size is fixed to be $\Delta t = 0.025$, so the number of coarse layers in discrete step is $20$. We can see relative $L^2$ error is decreasing as the number of training data increases for all loss functions. But the one using $Loss_1$ has smallest relative $L^2$ error if the number of training data in the discrete step is the same. For part (b), the number of training data $n_t$ is fixed to be $30$. We see the relative error is also decreasing as the number of coarse layers increases. Again, the best performance is the one with $Loss_1$.

\begin{figure}[h]
\centering
\subfloat[a][]{
\includegraphics[width=0.45\textwidth]{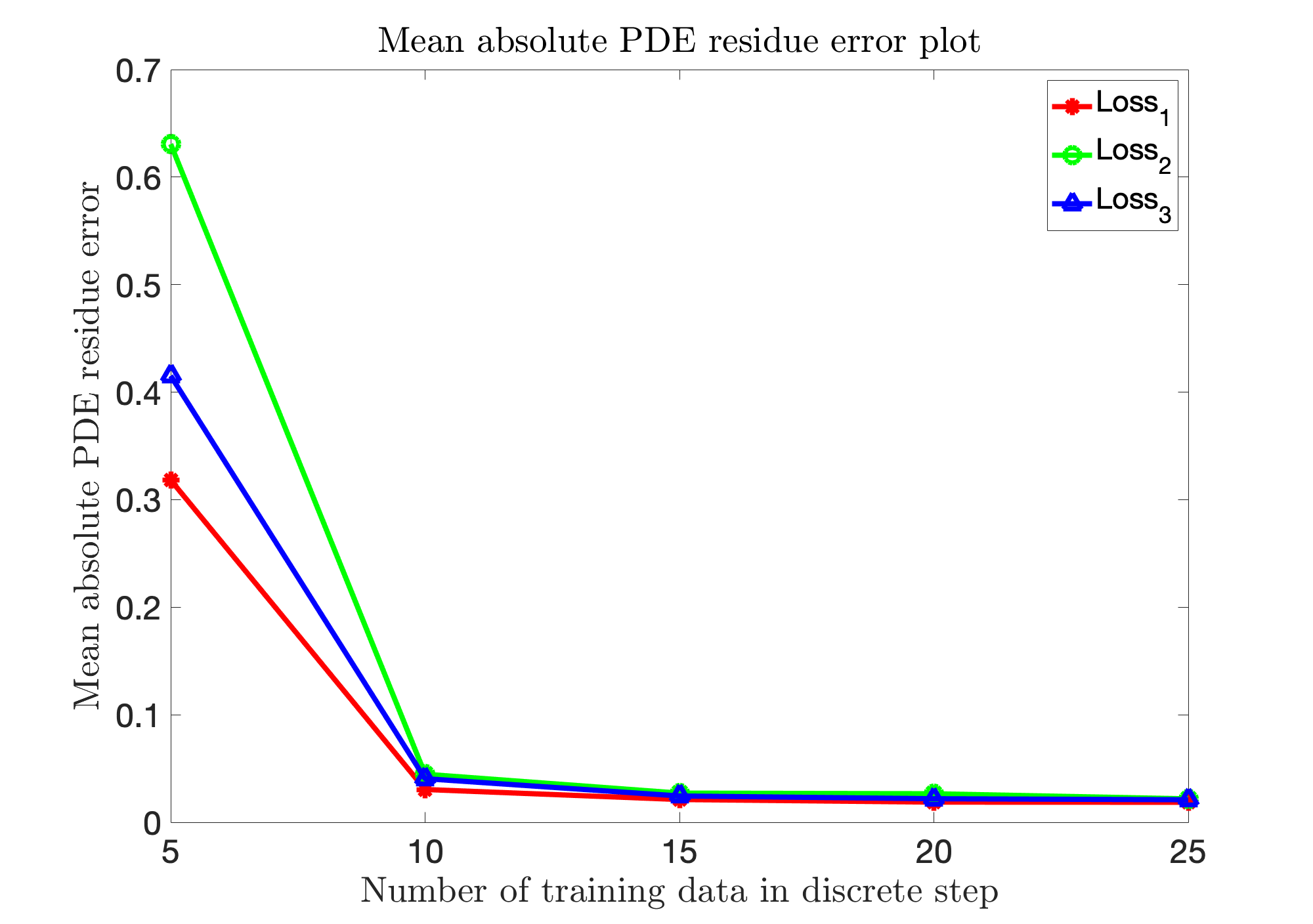}
\label{fig:ADV1d_PDEt}}
\qquad 
\subfloat[b][]{
\includegraphics[width=0.45\textwidth]{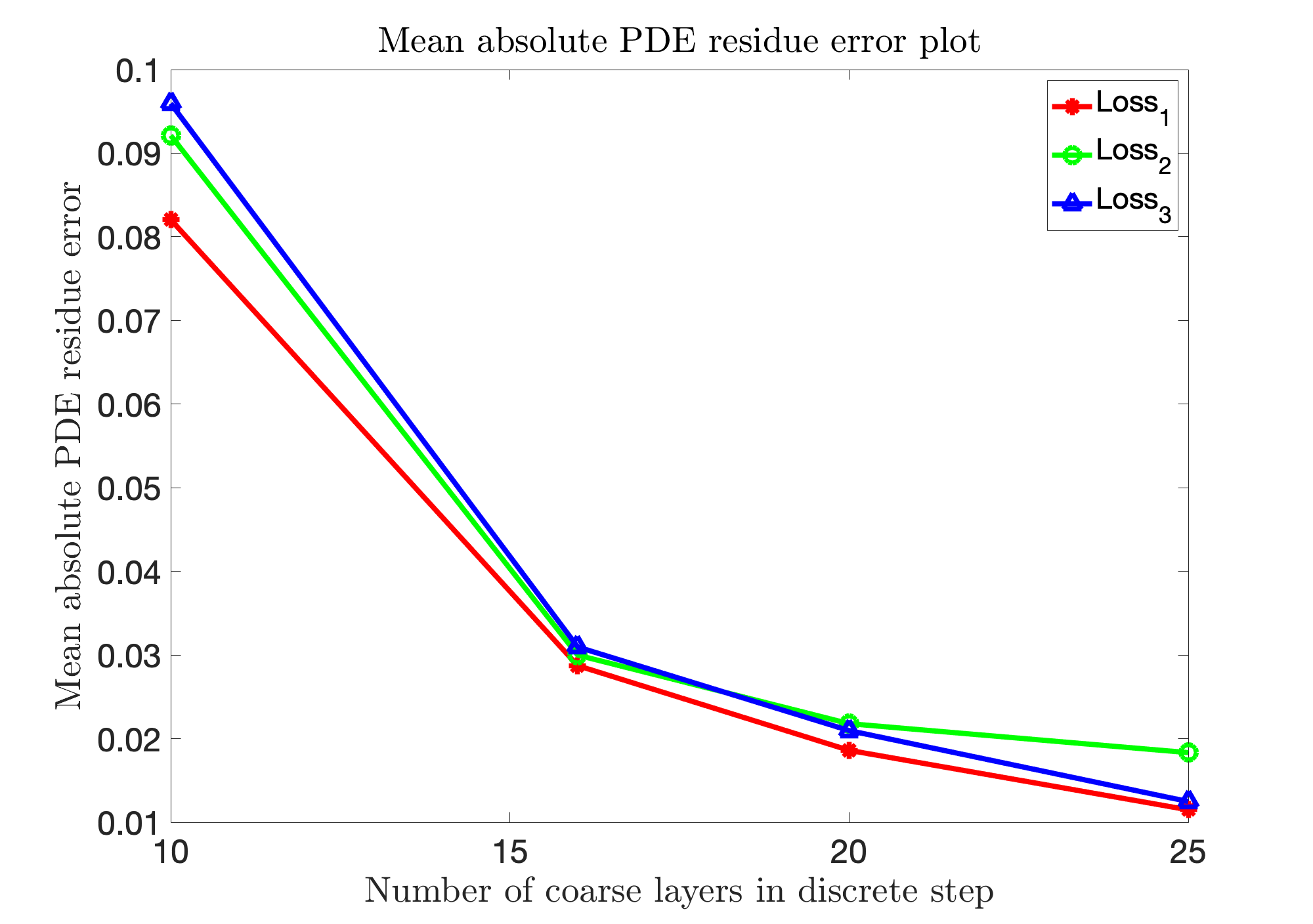}
\label{fig:ADV1d_PDEc}}
\caption{Advection equation's mean absolute PDE residue error plots using different loss functions. The number of collocation points is $30$ in the discrete step and $900$ in the continuous step. $300$ test points are randomly chosen in the whole domain $[0,0.5] \times [0,1]$. The red lines are the plots of Equation \ref{e13}. The green lines are the plots of  Equation \ref{e14} and the blue lines are the plots of Equation \ref{e15}. (a): mean absolute PDE residue error plots with different number of training data in discrete step. The number of coarse layers in discrete step is fixed to be $20$; (b): mean absolute PDE residue error plots with different number of coarse layers in discrete step. The number of training data in discrete step is fixed to be $25$.}
\label{fig:Adv1dPDE}
\end{figure}

Figure \ref{fig:Adv1dPDE} presents the results with the measure of mean absolute PDE residue error. The red lines are the plots using $Loss_1$. The green lines are the plots using $Loss_2$ and the blue lines are the plots using $Loss_3$. For part (a), the number of coarse layers in discrete step is fixe to be $20$. We can conclude that mean absolute PDE residue error is decreasing as the number of training data increases for all loss functions. But the one using $Loss_1$ has smallest relative $L^2$ error if the number of training data in the discrete step is the same. Also, as the number of training data increases, the differences between the mean absolute PDE residue error using the different loss functions become smaller. For part (b), the number of training data $n_t$ is fixed to be $30$. The mean absolute PDE residue error is decreasing as the number of coarse layers increases and the performance of using $Loss_1$ is the best.

\begin{figure}[h]
\centering
\subfloat[a][]{
\includegraphics[width=0.45\textwidth]{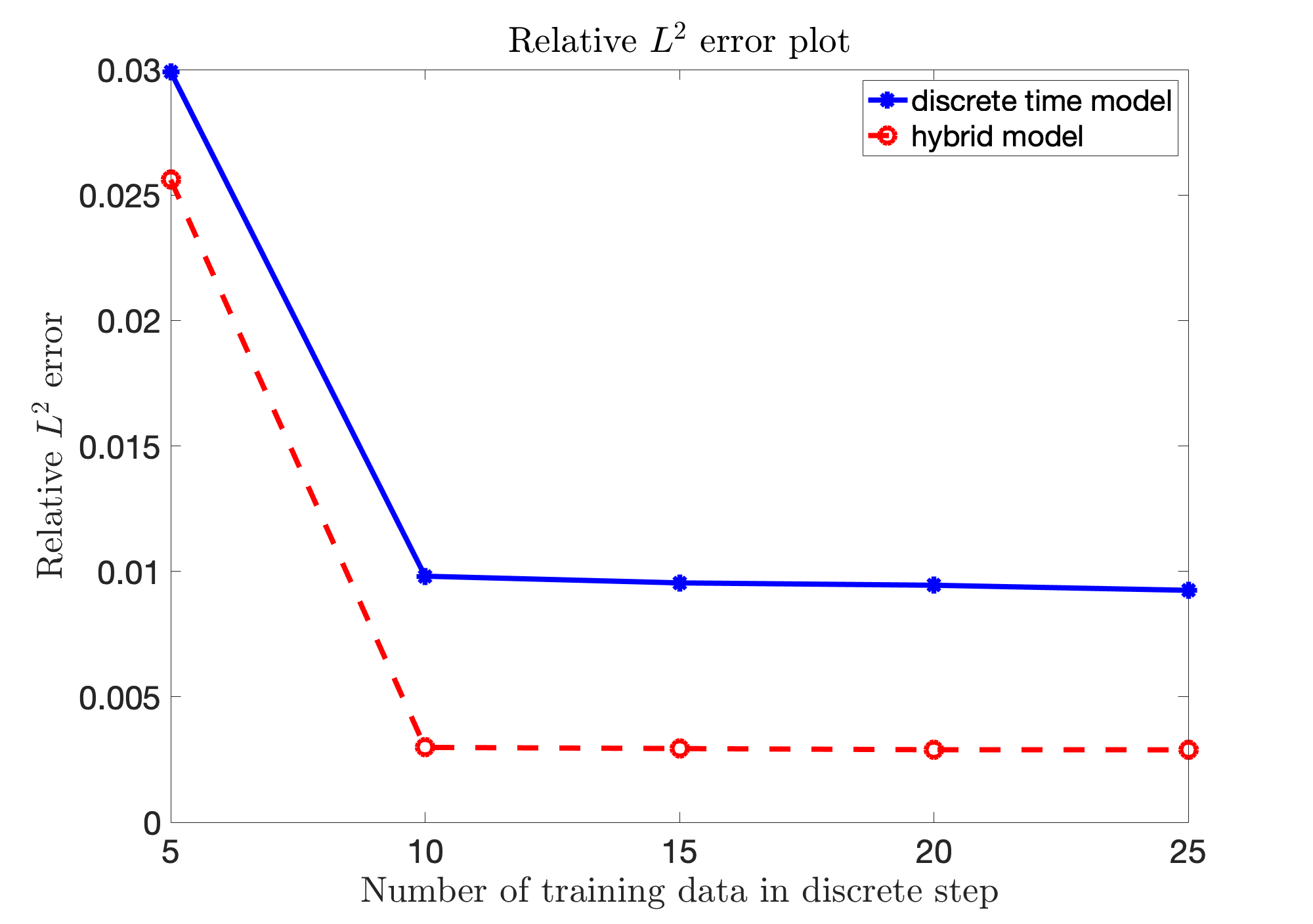}
\label{fig:ADV_tl2compare}}
\qquad 
\subfloat[b][]{
\includegraphics[width=0.45\textwidth]{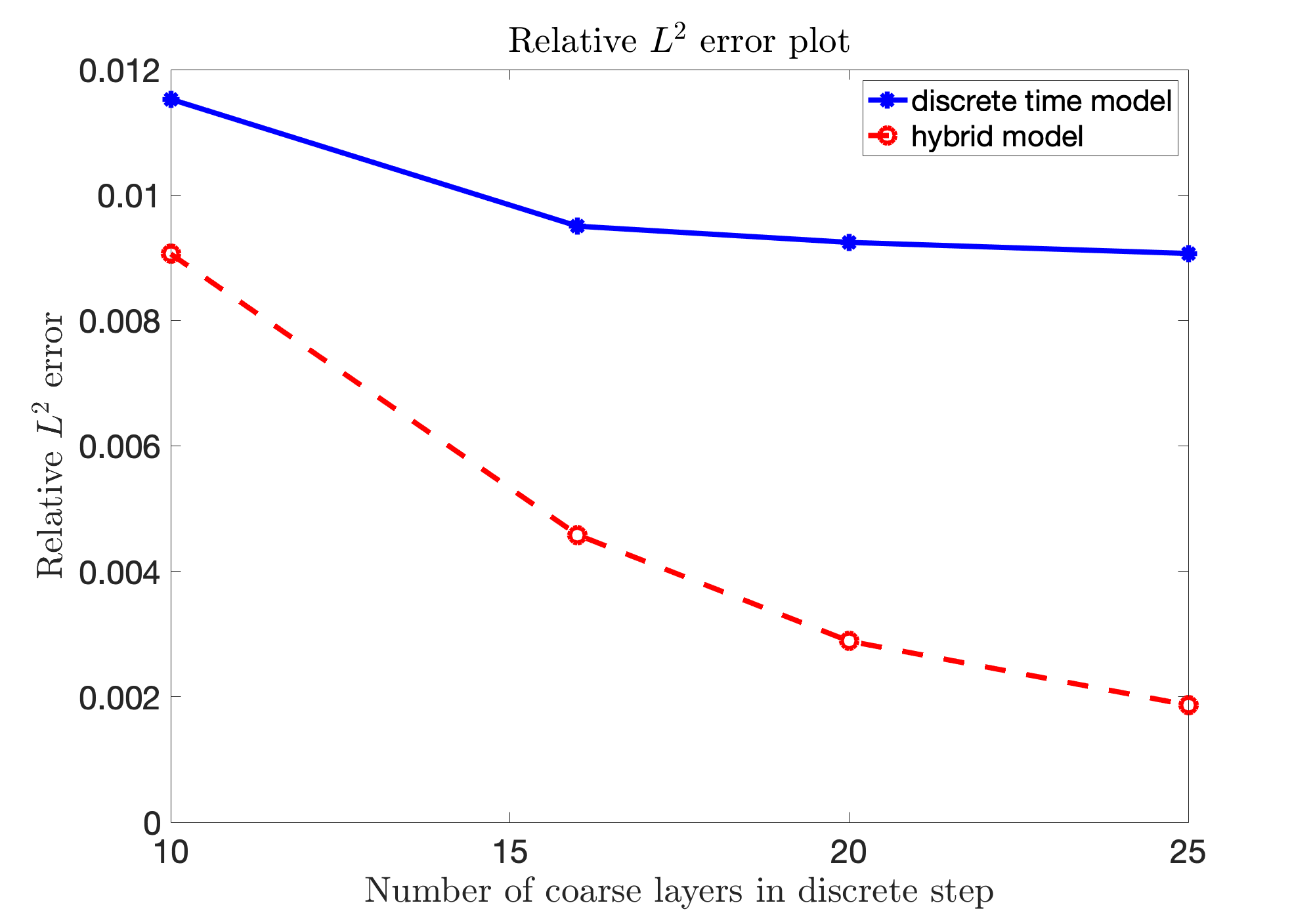}
\label{fig:ADV_cl2compare}}
\caption{Advection equation's relative $L^2$ error plots using loss function $Loss_1$. The number of collocation points is $30$ in the discrete step and $900$ in the continuous step. $300$ test points are randomly chosen in the whole domain $[0,0.5] \times [0,1]$. (a): relative $L^2$ error plots with different number of training data in discrete step. The number of coarse layers in discrete step is fixed to be $20$. The blue solid line is the plot of discrete time model and the red dashed line is the plot of hybrid model; (b): relative $L^2$ error plots with different number of coarse layers in discrete step. The number of training data in discrete step is fixed to be $25$. The blue solid line is the result of discrete time model and the red dashed line is the result of hybrid model.}
\label{fig:Adv1d_comparel2}
\end{figure}

We also apply the pure discrete time model for this forward problem in order to make a comparison with the hybrid model. The training, collocation and test data set used for discrete time model are the same as hybrid model. Figure \ref{fig:Adv1d_comparel2} presents the relative $L^2$ error of the two models using loss function $Loss_1$. The blue solid lines are from discrete time model and the red dashed lines are from hybrid model. From part (a), we can see the relative $L^2$ errors decrease as the number of training points increases for both discrete and hybrid models. This shows the convergence of the proposed method. With the same number of training points $n_t$, the relative $L^2$ errors of hybrid model are less than that of discrete model. From part (b), the relative $L^2$ errors decrease as the number of coarse layers increases for both models. But the hybrid model achieves smaller error compared to the discrete time model.

\begin{figure}[h]
\centering
\subfloat[a][]{
\includegraphics[width=0.45\textwidth]{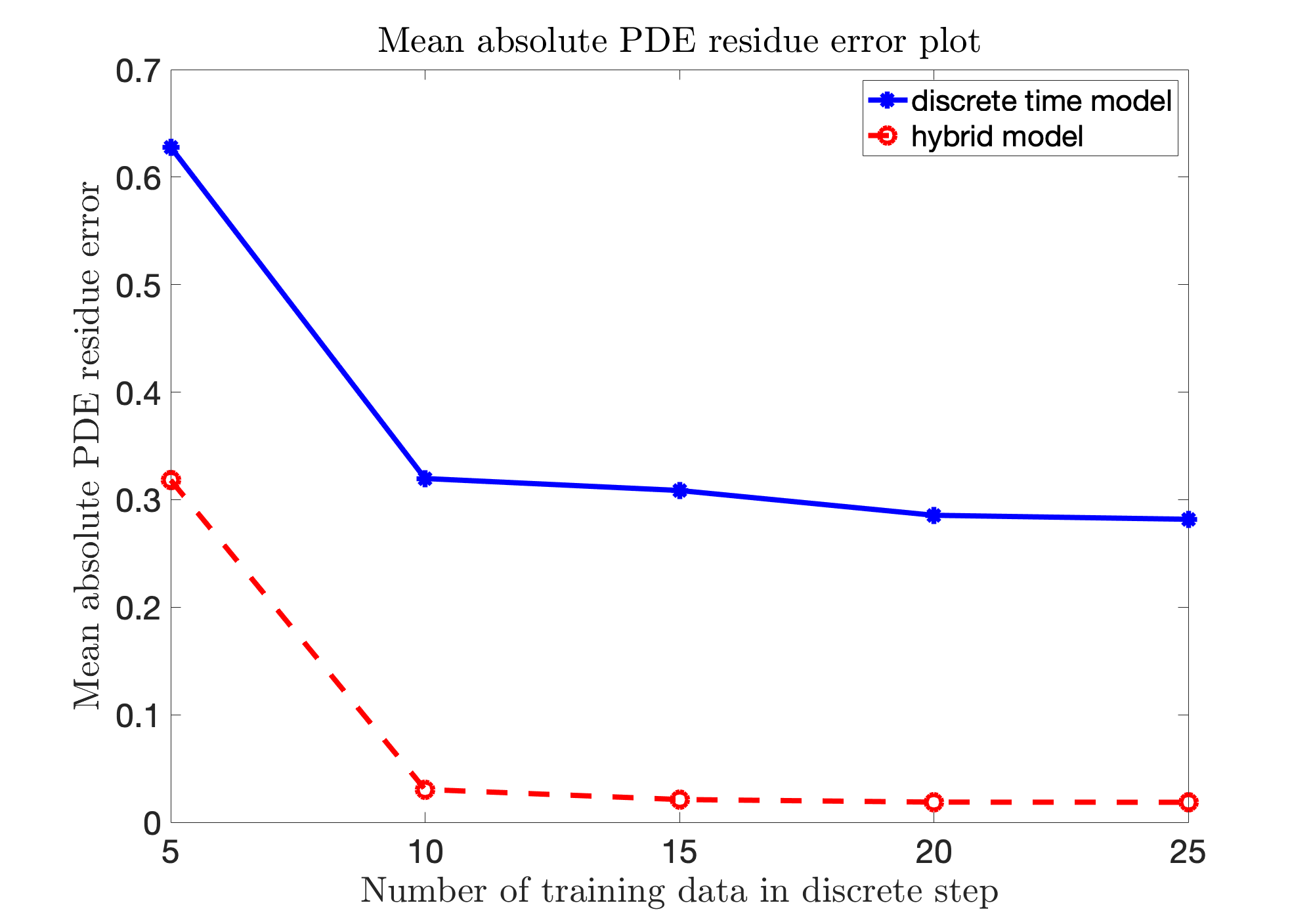}
\label{fig:ADV_tPDEcompare}}
\qquad 
\subfloat[b][]{
\includegraphics[width=0.45\textwidth]{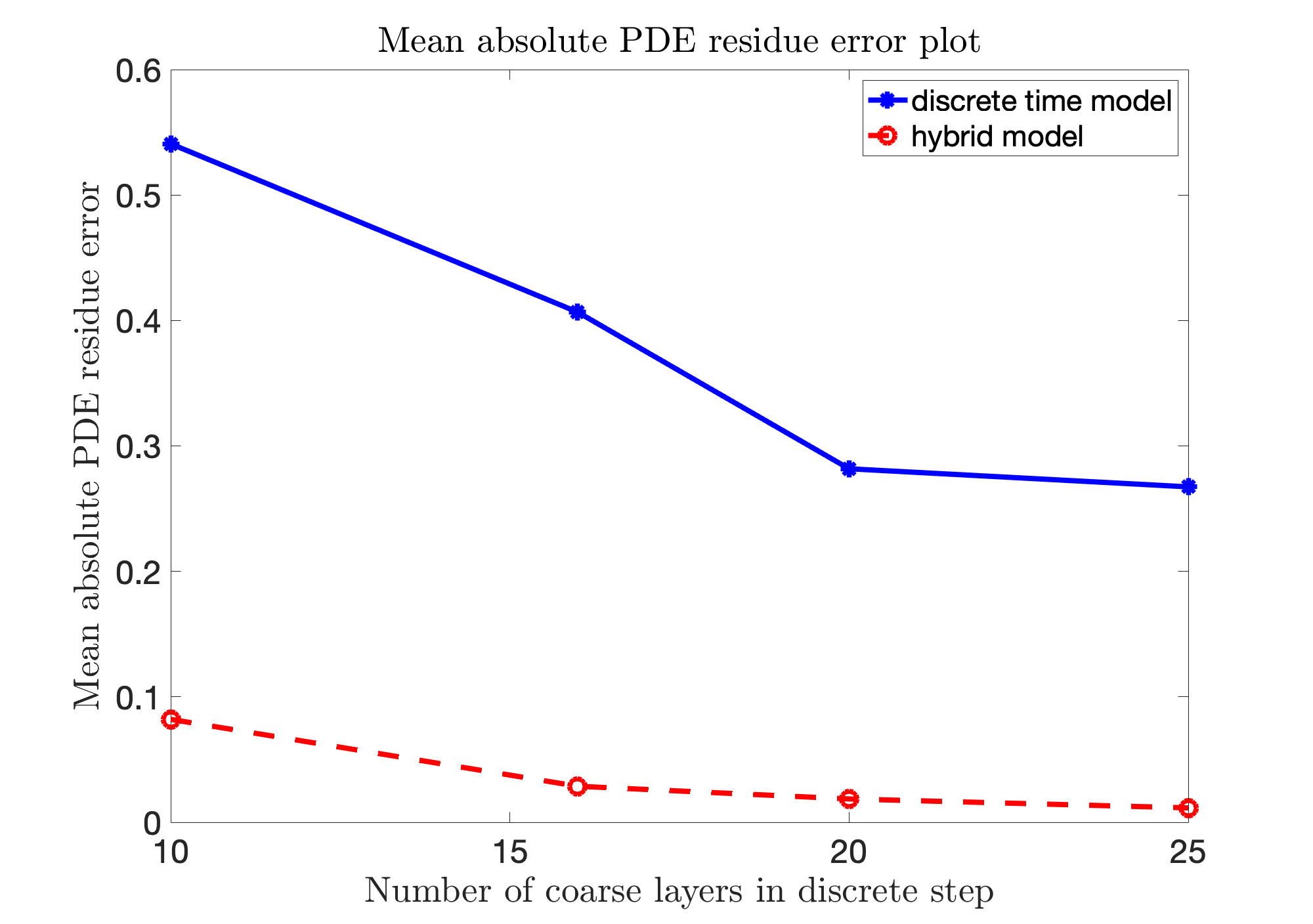}
\label{fig:ADV_cPDEcompare}}
\caption{Advection equation's mean absolute PDE residue error plots using loss function $Loss_1$. The number of collocation points is $30$ in the discrete step and $900$ in the continuous step. $300$ test points are randomly chosen in the whole domain $[0,0.5] \times [0,1]$. (a): mean absolute PDE residue error plots with different number of training data in discrete step. The number of coarse layers in discrete step is fixed to be $20$. The blue solid line is the plot of discrete time model and the red dashed line is the plot of hybrid model; (b): mean absolute PDE residue error plots with different number of coarse layers in discrete step. The number of training data in discrete step is fixed to be $25$. The blue solid line is the result of discrete time model and the red dashed line is the result of hybrid model.}
\label{fig:Adv1d_comparePDE}
\end{figure}

Figure \ref{fig:Adv1d_comparePDE} presents the mean absolute PDE residue error with loss function $Loss_1$. The blue solid lines are from discrete time model and the red dashed lines are from hybrid model. From part (a), we can see the mean absolute PDE residue errors decrease as the number of training points increases for models. What is more, the mean absolute PDE residue error of hybrid model are less than that of discrete model if the number of training points $n_t$ is the same. In part (b), the mean absolute PDE residue errors of discrete and hybrid models both have decreasing trends. However, mean absolute PDE residue errors of hybrid model are much less than that of discrete model if the number of coarse layers are the same. 

From Figures \ref{fig:Adv1d_comparel2} and \ref{fig:Adv1d_comparePDE}, we can conclude that hybrid model can effectively reduces the relative $L^2$ and PDE residue errors. Similar results can be obtained if we replace loss function $Loss_1$ with $Loss_2$ or $Loss_3$.
 
\subsection{One dimensional advection equation}
From the previous four numerical examples, we can see that different loss functions used in the training process indeed has an influence on the model performance. Our proposed three loss functions all consist of two terms in different scales. Thus the weight coefficients in Equations \ref{e13}, \ref{e14} and \ref{e15} need to be carefully adjusted in order for a meaning training process. Although we apply an adaptive procedure based on the relative magnitude of the two terms in loss functions, it might worse the model performance and cause stability issue in certain situation. We present one example in this section and modify loss function $Loss_2$ to deal with this problem. The model used here is the discrete time model and we only consider the forward problem to solve the following one dimensional advection equation,
\begin{equation}\label{e37}
    \frac{\partial}{\partial t}u + \frac{\partial}{\partial x}u =  0,
    \hspace{3mm} x \in [0,1], \hspace{3mm} t \in [0, 0.5]
\end{equation}
with initial condition,
\begin{equation}\label{e38}
    u(0, x) = 2000 sin(\pi x)
\end{equation}
and boundary conditions,
\begin{equation}\label{e39}
    u(t, 0) = -2000 sin(\pi t), \hspace{3mm}
    u(t, 1) = 2000 sin(\pi t)
\end{equation}
The true solution for this advection equation is,
\begin{equation}\label{e40}
    u(t, x) = 2000 sin(\pi (x - t))
\end{equation}

In order to investigate the impact of loss functions to the model performance, we gradually increase the number of training data $N_t$ and record the relative $L^2$ error on $100$ evenly spaced test points at $t = 0.5$. The step size in discrete time model is set to be $0.01$s and the time discretization scheme is the second order Adams-Bashforth method. The number of collocation points is fixed to be $N_c = 25$. The loss functions compared here is $Loss_1$, $Loss_2$, $Loss_3$ along with $Loss_4$ which is constructed from $Loss_2$. The following is the newly proposed loss function,
\begin{equation}\label{e41}
     Loss_4 = L_{\textit{LOO-MERE}} + \omega \textit{MERE}_r
\end{equation}
where 
\begin{equation}\label{e42}
    L_{\textit{LOO-MERE}} \displaystyle = \frac{1}{n}\sum_{i=1}^n (\frac{y_i - \mu_i}{y_i})^2
\end{equation}
represents the mean element-wise relative $L^2$ error and 
\begin{equation}\label{e43}
    \textit{MERE}_r = \frac{1}{N_c}\sum_{i} (\frac{ (u_t(\bold{x}_c, t_c) - \mathcal{T}_{ \bold{x}}^{\lambda} u(\bold{x}_c, t_c) )_i}{(\mathcal{T}_{ \bold{x}}^{\lambda} u(\bold{x}_c, t_c))_i})^2
\end{equation}
represent the mean element-wise relative PDE residue error. In Equations \ref{e42} and \ref{e43}, the denominator in each summation term can be zero in certain situation. In this case, a small positive number is added to the denominator to avoid computational issues. Another way to deal with this problem is to translate the PDE solution $u$ so that all denominators in the summation terms are positive.

In this way, the two terms in $Loss_4$ are both normalized to the same scale. So the training process will be more stable and the adaptive procedure to adjust weight will be much easier during the training process. All loss functions are minimized using BFGS algorithm. The initial weight coefficients are chosen to be $\omega = 1$ for all loss functions. The rate factors are $1.2$ for $Loss_1$, $Loss_2$ and $Loss_4$, and $0.6$ for $Loss_3$. The maximal iteration number for updating the weight coefficients is $5$.

\begin{figure}[h]
\centering
\subfloat[a][]{
\includegraphics[width=0.45\textwidth]{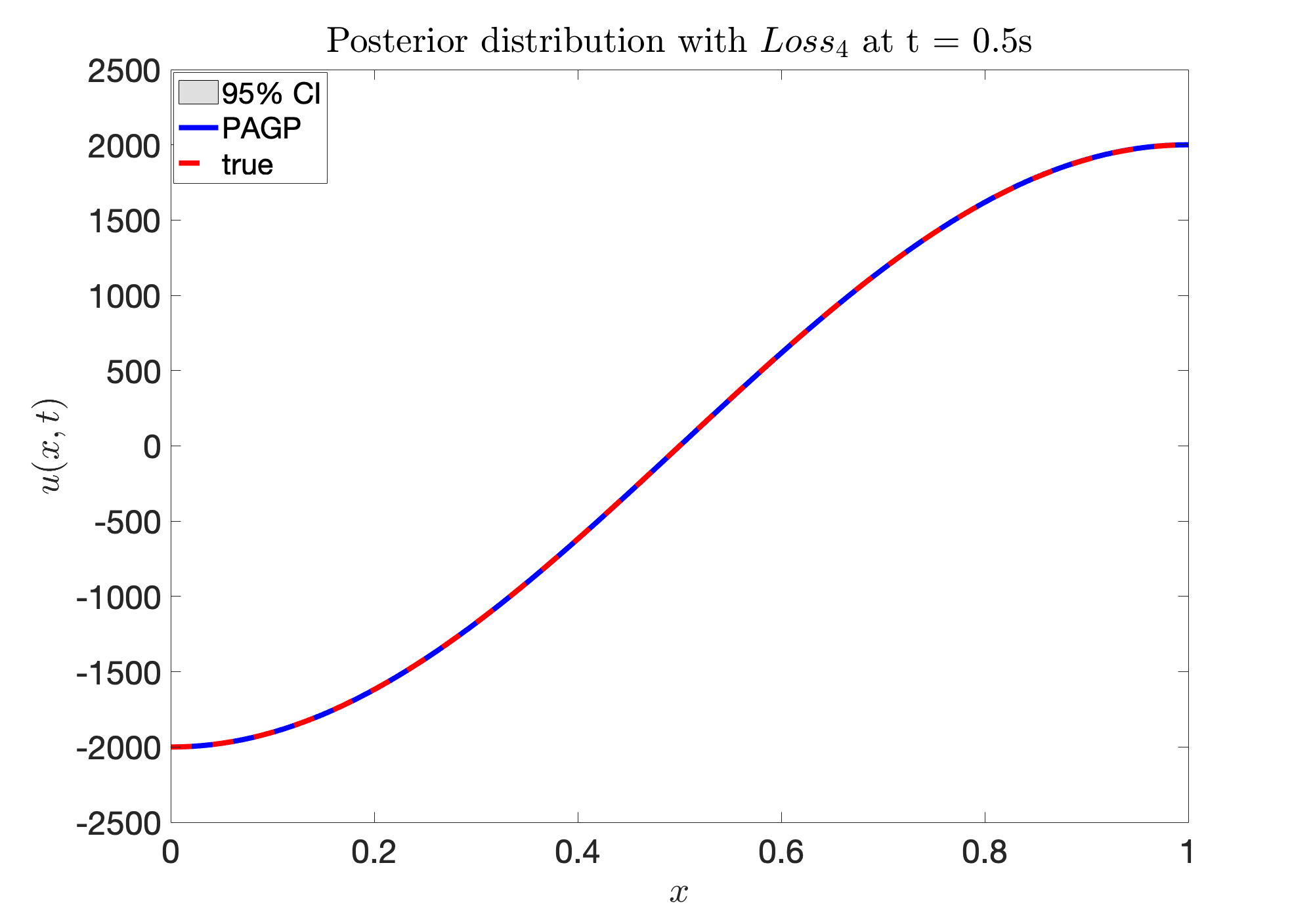}
\label{fig:ADV_loss4pred}}
\qquad 
\subfloat[b][]{
\includegraphics[width=0.45\textwidth]{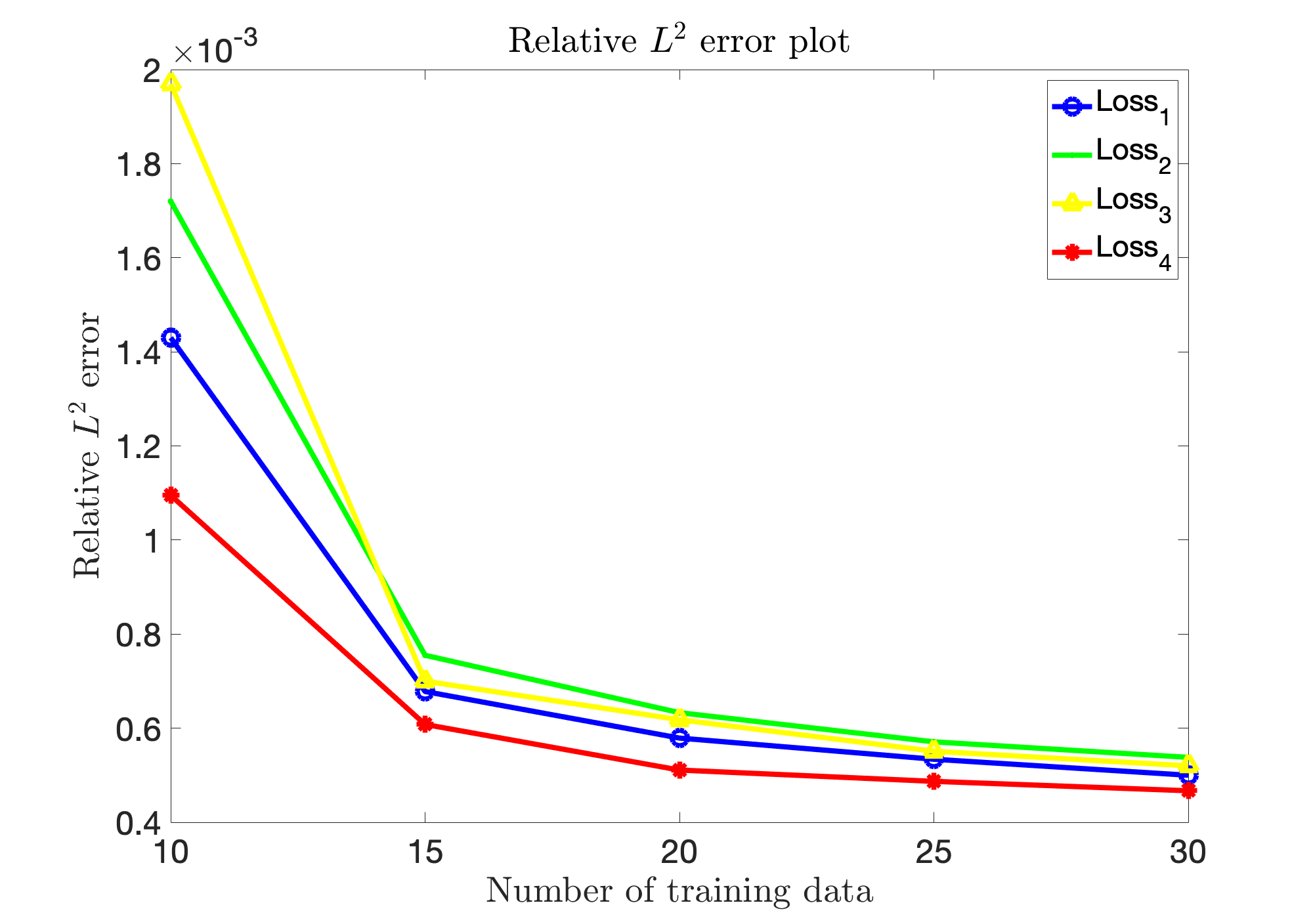}
\label{fig:ADV_4l2compare}}
\caption{Equation \ref{e37}'s posterior distribution plot and relative $L^2$ error plot using different loss functions. The time discretization scheme is second order Adams-Bashforth and the step size is set to be $0.01$s. The number of collocation points is fixed to be $N_c = 25$. There are $100$ evenly spaced test points at $t = 0.5$. (a): posterior distribution plot using $Loss_4$ at $t=0.5$s. The number of training points is $N_t = 25$. (b): relative $L^2$ error plots along with increasing number of training data $N_t = [10, 15, 20, 25, 30]$ corresponding to four different loss functions $Loss_1$, $Loss_2$, $Loss_3$ and $Loss_4$.}
\label{fig:ADV_4l2}
\end{figure}

Figure \ref{fig:ADV_loss4pred} presents the posterior distribution plot at $t =0.5$s using $Loss_4$. The number of training data is $N_t = 25$. The time step size is $0.01$s. The number of collocation data is $N_c = 25$. The blue solid line is the PAGP plot and the red dashed line is the true solution. The grey area represents the $95\%$ confidence interval, i.e. plus/minus two standard deviation around the posterior mean. Part (b) depicts the relative $L^2$ error plots of different loss functions with increasing number of training points $N_t$. We can see the red line corresponding to $Loss_4$ achieves the best performance. In this experiment, the function values at different spatial locations change dramatically and the second term in loss functions $Loss_1$, $Loss_2$ and $Loss_3$ is much bigger than the first term. This makes the weight tuning process difficult and might cause the training process unstable. The plots in Figure \ref{fig:ADV_4l2compare} is the best performance results of all loss functions after tuning the weight coefficient in corresponding loss functions. We found that using $Loss_4$ can make above process much easier and stable due to the normalization pre-process of the two terms in this loss function. Thus we recommend using this loss function in a similar problem setting.

\section{Conclusion}
\label{Summary}
In this paper, a new Gaussian process regression framework incorporated with physical laws is established to solve the forward and inverse problems of PDEs. Three different models are developed and four types of PAGP loss functions are constructed and discussed. An adaptive weight updating procedure is adopted to assisted the training process. Five numerical examples are presented in order to illustrate the performance of the proposed PAGP models. The continuous time model treats the temporal domain the same as spatial domain. So it is more concise and flexible. For discrete and hybrid methods, a Bayesian active learning scheme can be involved to enhance the accuracy and reduce the variance. The hybrid model combines the merits of continuous and discrete time models. And it can effectively reduce the PDE residue error. We recommend the hybrid model with loss function $Loss_1$ but one should make the selection based on problems at hand. For instance, one should applies $Loss_4$ in the problem similar to the last example in numerical section.

With our PAGP models, one can solve a given PDE with boundary and initial conditions or discovering unknown coefficients accurately. The building block for PAGP is Gaussian process which is the same as PIGP. But instead of incorporating physical information into GP through complex corvariance functions, we directly add PDE constrains in GP loss functions. So the framework is more simple and flexible. Unlike PINN which is based on neural network, the probabilistic workflow of GP provide a natural way for PAGP models to quantify the uncertainties. Also, there are less parameters to tune compare to PINN models. Thus, the training process can be more robust and the model calibration can be easier. 

In the future, we plan to apply the PAGP models to study more complex physical systems and find an effective way to reduce the model uncertainties resulting from noises in the data. For continuous time model, the posterior distribution variance need to be reduce efficiently. The time discretization scheme used in discrete and hybrid models in this paper is the second order Adams-Bashforth method. Other discretization scheme such as Runge–Kutta method can also be employed similarly but with some efforts. In particular, investigating the impact of different time discretization scheme to the performance of discrete and hybrid PAGP models can be interesting. Other potential future works include investigating theoretical concepts like prior consistency and posterior robustness for our models.

\section*{Acknowledgement}
We gratefully acknowledge the support from the National Science Foundation (DMS-1555072, DMS-1736364, and DMS-2053746), and Brookhaven National Laboratory Subcontract 382247, and U.S. Department of Energy (DOE) Office of Science Advanced Scientific Computing Research program DE-SC0021142.

\bibliographystyle{abbrv}
\bibliography{mybib}
\nocite{bhouri2021gaussian, xiong2020clustered, shi2011gaussian, wang2021explicit, raissi2018hidden, yuan2020modeling, tran2020multi, peherstorfer2018survey, lam2020multifidelity, wang2020physics, zhu2019physics, yang2019physics, wang2020and, karumuri2020simulator, kharazmi2021hp, hennig2015probabilistic, leveque2007finite, conrad2017statistical, murphy2012machine, tihonov1963solution, butcher2016numerical, rasmussen2001occam, hensman2013gaussian, MR3996296, MR2650669, iserles2009first, hartikainen2010kalman, owhadi2015brittleness, schober2014probabilistic, stuart2018posterior, basdevant1986spectral, rudy2017data, wang2017comprehensive, stein1987large, brunton2016discovering}
    
\end{document}